\documentclass[journal]{IEEEtran}
\usepackage[T1]{fontenc}
\usepackage{cite}
\usepackage{amsmath,amssymb,amsfonts}
\usepackage{algorithmic}
\usepackage{graphicx}
\usepackage{url}
\usepackage{textcomp}
\usepackage[table]{xcolor}
\usepackage{multirow}
\usepackage{bm}
\usepackage{float}
\usepackage{algorithm}
\usepackage[stretch=30,shrink=30]{microtype}
\usepackage[hidelinks]{hyperref}
\hypersetup{
  pdftitle={Event-Driven Refresh and Recurrence Memory to Reduce Stale Grounding in Referring Video Object Segmentation},
  pdfauthor={Abu Hanif Muhammad Syarubany, Jaehyun Jang, Siwoo Lim, Seungyeon Ryu, Chang D. Yoo},
  pdfkeywords={Referring video object segmentation (RVOS), multimodal large language models (MLLMs), Segment Anything Model (SAM), SAM2, Sa2VA, temporal grounding, adaptive sampling}
}

\newcommand{\hl}[1]{#1}
\newcommand{\palegreen}{\cellcolor{green!10}}

\makeatletter
\long\def\@makecaption#1#2{%
  \ifx\@captype\@IEEEtablestring
    \footnotesize\bgroup\par
    \setbox\@tempboxa\hbox{\normalfont\footnotesize\textbf{#1.}\nobreakspace\nobreakspace #2}%
    \ifdim\wd\@tempboxa>\hsize\else\centering\fi
    \@IEEEtabletopskipstrut{\normalfont\footnotesize\textbf{#1.}\nobreakspace\nobreakspace #2}\par
    \addvspace{0.5\baselineskip}\egroup
    \@IEEEtablecaptionsepspace
  \else
    \@IEEEfigurecaptionsepspace
    \setbox\@tempboxa\hbox{\normalfont\footnotesize\textbf{#1.}\nobreakspace\nobreakspace #2}%
    \ifdim\wd\@tempboxa>\hsize
      \setbox\@tempboxa\hbox{\normalfont\footnotesize\textbf{#1.}\nobreakspace\nobreakspace}%
      \parbox[t]{\hsize}{\normalfont\footnotesize\noindent\unhbox\@tempboxa #2}%
    \else
      \hbox to\hsize{\normalfont\footnotesize\box\@tempboxa\hfil}%
    \fi
  \fi}
\makeatother

\newcommand{\runtfix}{\parfillskip=0pt plus 0.6\linewidth\relax}
\newcommand{\runtoff}{\parfillskip=0pt plus 1fil\relax}
\begin{document}
\flushbottom

\title{Event-Driven Refresh and Recurrence Memory to Reduce Stale Grounding in Referring Video Object Segmentation}

\author{Abu Hanif Muhammad Syarubany,
        Jaehyun Jang,
        Siwoo Lim,
        Seungyeon Ryu,
        and~Chang D. Yoo\\
        {\small\itshape Korea Advanced Institute of Science \& Technology (KAIST)}%
\thanks{This work has been submitted to the IEEE for possible publication. Copyright may be transferred without notice, after which this version may no longer be accessible.}%
\thanks{The authors are with the School of Electrical Engineering, Korea Advanced Institute of Science and Technology (KAIST), Daejeon 34141, Republic of Korea. Corresponding author: Chang D. Yoo (e-mail: cd\_yoo@kaist.ac.kr).}}

\markboth{Preprint}{Syarubany \MakeLowercase{\textit{et al.}}: Event-Driven Refresh and Recurrence Memory for Referring Video Object Segmentation}

\maketitle

\begin{abstract}
Referring Video Object Segmentation (RVOS) aims to produce a pixel-accurate mask sequence for an object specified by natural language. Sa2VA combines a multimodal large language model with SAM2 for grounded segmentation; however, its inference typically grounds the query from a small fixed set of initial keyframes and then relies on propagation. In long or dynamic videos, this can cause stale grounding and persistent false positives when the object composition changes (e.g., distractors enter or the target disappears/re-appears). 
We propose \emph{Event-Driven Refresh + Recurrence Memory} (EDRRM), an enhancement that selectively re-invokes Sa2VA only at stable change points. EDRRM triggers refresh boundaries using an EMA-smoothed event score computed from tracking-derived cues (births/deaths and coarse composition/layout changes) with temporal constraints. A recurrence memory further retrieves anchor frames via CLIP similarity to re-condition the model on re-appearance events. 
\hl{Experiments on Ref-DAVIS17, MeViS, and ReVOS show that EDRRM achieves a competitive accuracy-efficiency trade-off relative to fixed-window and FrameDiff-SSIM baselines, maintaining comparable or superior J\&F scores at substantially lower average refresh-call budgets and reducing false-positive failures. End-to-end runtime analysis further confirms that the overhead introduced by tracking, CLIP-based recurrence matching, and the identifiability gate remains modest relative to the dominant Sa2VA inference cost, thereby validating the efficiency of the proposed pipeline.}
\end{abstract}


\section{Introduction}
\label{sec:intro}

\IEEEPARstart{R}{eferring} Video Object Segmentation (RVOS) aims to output a binary mask sequence
$\{ \mathbf{M}_t \}_{t=1}^{T}$ for a target described by a natural-language query.
Compared with conventional video object segmentation (VOS), RVOS replaces mask prompts with
language, enabling more natural user interaction but introducing additional ambiguity under
occlusion, viewpoint change, and multi-instance clutter.
Consequently, RVOS progress is commonly measured on benchmarks such as Ref-DAVIS17~\cite{davis}, MeViS~\cite{mevis}, and ReVOS~\cite{revos}.

Recent advances have combined foundation segmentation with multimodal large language models (MLLMs),
enabling densely grounded understanding in videos. In particular, Sa2VA couples an MLLM (LLaVA-style)
with SAM2 to produce grounded \texttt{[SEG]} tokens and high-quality masks \cite{sa2va2025,sam2_2024,liu2023visual,sam_2023}.
However, Sa2VA’s inference relies on a \emph{small fixed set of keyframes} to establish grounding, which uses the \emph{first five frames} of the input video as keyframes for prompting and then
propagates masks through the remaining frames \cite{sa2va2025}.
In long or dynamic videos, the initial grounding can become stale when the object composition changes (e.g., target disappears/re-appears, new distractors enter, or layout shifts), often manifesting as false positives or drift.

Motivated by this limitation, we study the following core problem: \emph{how to refresh grounded segmentation only when the video content meaningfully changes, without incurring the cost of dense re-prompting.} We propose an \textbf{Event-Driven Refresh + Recurrence Memory (EDRRM)} enhancement to Sa2VA~\cite{sa2va2025} (Figs.~\ref{fig:process_flow} and~\ref{fig:recurrence_flow}). Our method monitors the video stream with an object tracker and simple change cues (e.g., instance births/deaths and coarse layout shifts) to detect stable change points and trigger a re-invocation of Sa2VA on a compact frame batch around the event. In addition, we introduce a recurrence memory that stores embeddings of recently disappeared instances and matches them upon re-appearance using CLIP similarity, enabling anchor-frame retrieval to re-condition Sa2VA and reduce identity switches \cite{radford2021clip,li2024masa}.

We evaluate on RVOS benchmarks including Ref-DAVIS17~\cite{davis,ponttuset2017davis}, MeViS~\cite{mevis}, and ReVOS~\cite{revos}, and compare against two practical
sampling baselines: (i) \emph{fixed-window} re-prompting and (ii) \emph{FrameDiff-SSIM} change detection
based on structural similarity \cite{ssim2004}.
Overall, our approach keeps Sa2VA’s strong grounded segmentation backbone while reducing stale-grounding failures
by re-invoking Sa2VA \emph{only} at event-driven change points and recurrence-driven re-appearances.
\hl{Crucially, EDRRM frames re-grounding as a \emph{scheduling and control problem}: rather than applying a fixed inference, it explicitly decides \emph{when} to invoke grounding based on object-composition change signals, cleanly separating the scheduling mechanism from the segmentation backbone itself.}

\section{Related Work}
\label{sec:related_work}

\textbf{RVOS models and benchmarks.}
RVOS requires segmenting the instance referred by a natural-language expression across time, which demands both correct grounding and robust temporal association under occlusion, distractors, and motion-centric cues. Representative CNN/transformer-based RVOS methods perform multimodal interaction between language and video features to improve temporal reasoning and instance discrimination~\cite{seo2020urvos,refvos2020,referformer2022,mttr2021,soc2023,gavrilyuk2018actor}. Efficiency-oriented settings have also been explored via online/semi-online inference to support streaming with competitive accuracy~\cite{onlinerefer2023}. Progress is driven by diverse benchmarks covering static and motion-heavy expressions, including Ref-DAVIS17~\cite{davis}, MeViS~\cite{mevis}, and ReVOS~\cite{revos}, which emphasize motion expressions and highlights temporal grounding failures.

\textbf{Foundation models and grounded segmentation with MLLMs.}
Promptable foundation segmentation (SAM) and its video extension (SAM2) enable strong mask prediction with minimal supervision~\cite{sam_2023,sam2_2024}, while MLLMs such as LLaVA improve instruction-following visual understanding~\cite{liu2023visual}. Sa2VA integrates SAM2 with an LLaVA-style MLLM to produce grounded \texttt{[SEG]} tokens for dense understanding in images and videos~\cite{sa2va2025}. However, Sa2VA-style pipelines typically ground the target from a compact set of keyframes and then rely on propagation, which can become brittle when video content evolves (e.g., target disappearance/re-appearance, distractor entry, or layout shifts), motivating selective re-grounding rather than purely relying on long-range propagation.

\textbf{Temporal memory, association, and robustness.}
Long-term VOS literature shows that explicit memory and association are critical for stable propagation in long videos~\cite{cheng2021stcn,cheng2022xmem,yang2021aot}. In parallel, retrieval-style re-identification using shared embedding spaces (e.g., CLIP) supports matching instances across time~\cite{radford2021clip}, and tracking/association modules such as MASA provide strong temporal continuity cues~\cite{li2024masa}. Robust RVOS further studies semantic mismatch cases where the query target is absent or ambiguous, motivating gating/consensus beyond per-frame confidence~\cite{li2023r2vos}. Our work draws from these directions: instead of refreshing at fixed intervals or low-level frame-difference heuristics, we propose event-driven re-grounding with recurrence-aware anchors to reduce stale-grounding drift in Sa2VA-style inference.

\hl{\textbf{Adaptive inference and temporal decision policies.} Beyond fixed-schedule inference, a growing line of work frames temporal re-inference as a scheduling or control problem. Adaptive frame sampling methods for video recognition selectively process only informative frames based on content-driven cues}~\cite{wu2019adaframe,ghodrati2021frameexit}\hl{, reducing redundant computation while preserving accuracy. Confidence-based early-exit frameworks defer computation to only those inputs that require deeper processing~\cite{ghodrati2021frameexit}. These methods share the core insight of EDRRM: rather than applying a uniform inference schedule, the computation is conditioned on the input signal. Our work extends this idea to RVOS by treating re-grounding as an event-triggered scheduling decision, where object-composition changes derived from the tracking stream determine when Sa2VA is re-invoked, rather than relying on fixed windows or low-level appearance differences.}

\section{Methodology}
\label{sec:method}

\begin{figure*}[t]
  \centering
  \includegraphics[width=\textwidth]{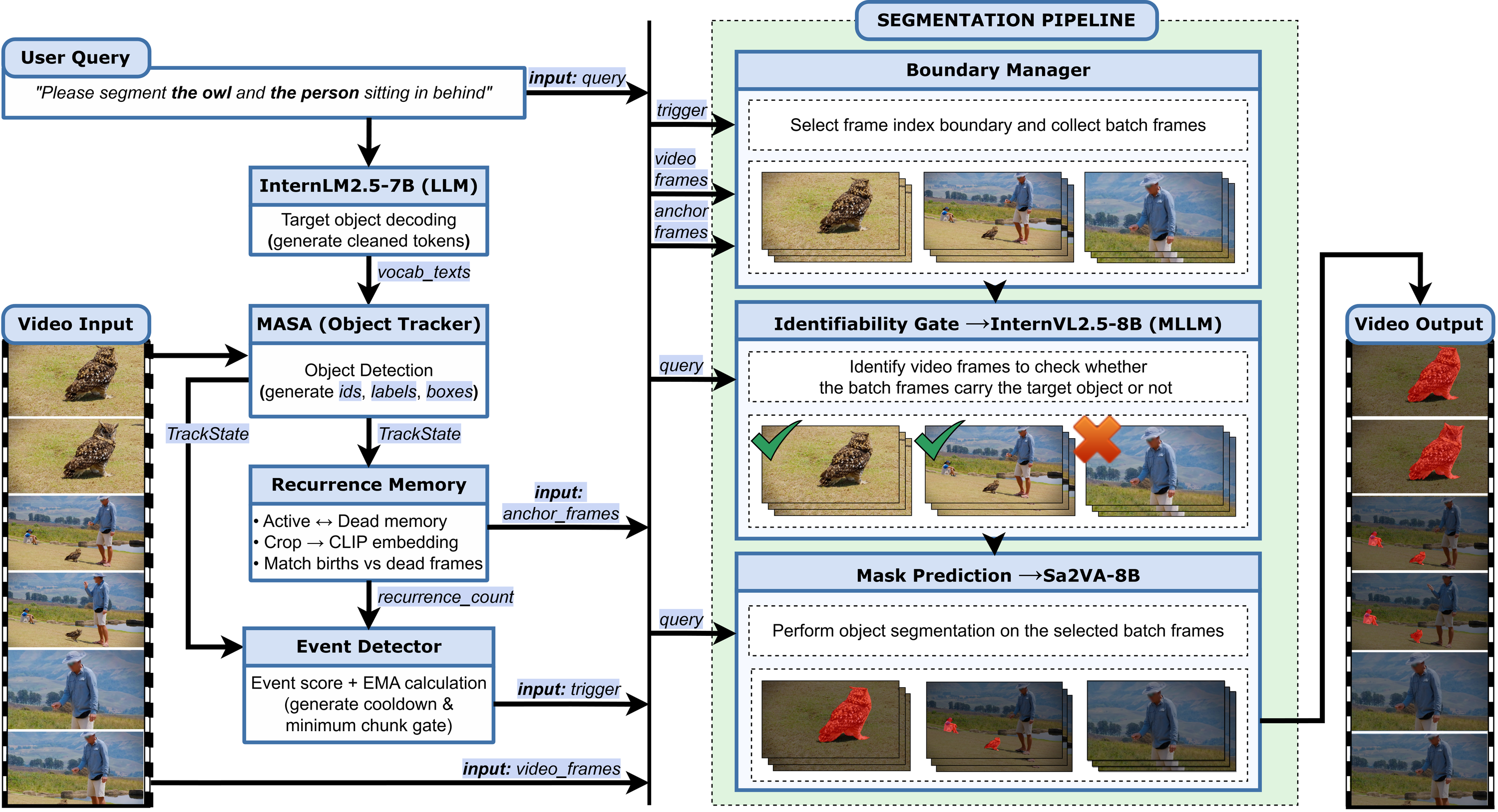}
  \caption{\textbf{Process flow of the proposed pipeline.}}
  \label{fig:process_flow}
\end{figure*}

\begin{figure*}[t]
  \centering
  \includegraphics[width=\textwidth]{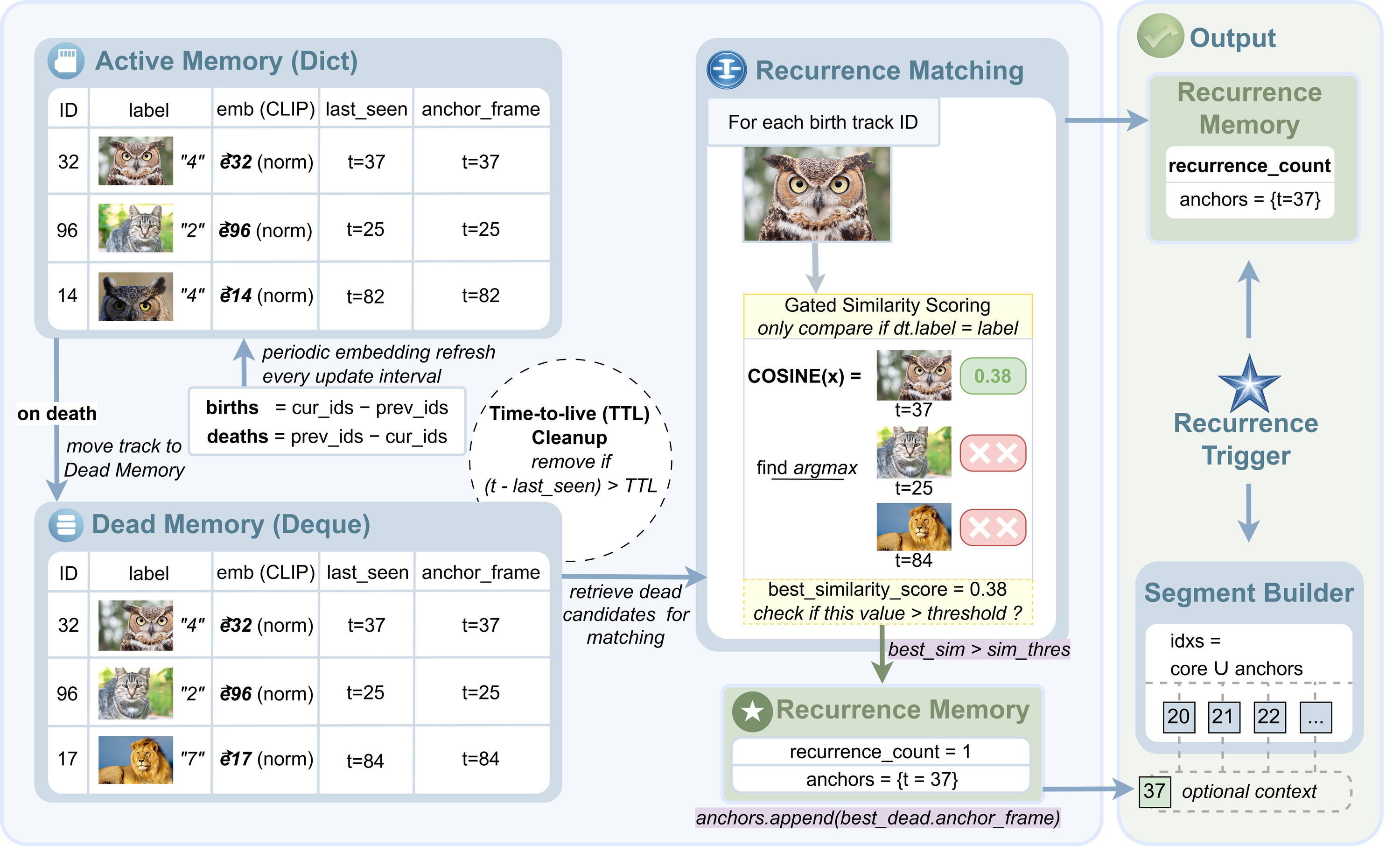}
  \caption{\textbf{Recurrence memory flowchart.}}
  \label{fig:recurrence_flow}
\end{figure*}

Fig.~\ref{fig:process_flow} illustrates the end-to-end process flow of our event-driven referring video object segmentation (RVOS) system, and Fig.~\ref{fig:recurrence_flow} details the proposed recurrence memory mechanism. Our goal is to reduce redundant segmentation calls and improve temporal robustness by triggering Sa2VA~\cite{sa2va2025} segmentation only when object-level changes (events) are detected, while additionally handling re-appearance (recurrence) via CLIP-based matching.

\subsection{Problem Setup and Notation}
\label{subsec:problem_setup}
Let a video be a sequence of RGB frames $\{\mathbf{I}_t\}_{t=0}^{T-1}$ and a user referring query be $q$ (e.g., ``segment the owl and the person''). Our objective is to produce a binary mask sequence $\{\mathbf{M}_t\}_{t=0}^{T-1}$ where $\mathbf{M}_t \in \{0,1\}^{H\times W}$ indicates the target object(s) in frame $t$.

We use an object tracker to obtain a per-frame track state:
\begin{equation}
\mathcal{S}_t = \{(id_i^t,\ \ell_i^t,\ \mathbf{b}_i^t)\}_{i=1}^{N_t},
\end{equation}
\looseness=-1 where $id_i^t$ is the track identity, $\ell_i^t$ is the (open-vocabulary) class label, and $\mathbf{b}_i^t=(x_1,y_1,x_2,y_2)$ is an axis-aligned bounding box.

\subsection{System Overview}
\label{subsec:overview}
As shown in Fig.~\ref{fig:process_flow}, the pipeline consists of: (i) \textbf{query decoding} to produce cleaned tokens for tracking and segmentation, (ii) \textbf{tracking-driven event scoring} to decide segmentation boundaries, (iii) \textbf{recurrence memory} to detect re-appearances and retrieve anchor frames (Fig.~\ref{fig:recurrence_flow}), (iv) \textbf{boundary manager} to assemble a frame batch for Sa2VA, (v) \textbf{identifiability gate} to prevent hallucinated segmentation, and (vi) \textbf{mask prediction} with Sa2VA on selected frames only. Finally, masks are re-aligned back to the original timeline.

\textbf{Implementation note.}
In our implementation, the query-decoding LLM (InternLM2.5-7B) and the identifiability MLLM (InternVL2.5-8B) are both instruction-following components already packaged within the Sa2VA-8B release. Thus, EDRRM does not require the deployment of additional large models beyond Sa2VA-8B; it only changes \emph{when} Sa2VA is invoked and how input frame batches are constructed.

\subsection{Query Decoding for Tracking and Segmentation}
\label{subsec:query_decoding}
\looseness=-1 Given $q$, an LLM (InternLM2.5-7B in our implementation) generates a cleaned token set $\mathcal{V}$ (``vocab\_texts'') and a segmentation prompt $p$ used by the RVOS model (Sa2VA). The tokens $\mathcal{V}$ are passed to the tracker to improve detection/association for target-relevant objects, while $p$ is used to drive segmentation. This stage corresponds to the ``Target object decoding'' block in Fig.~\ref{fig:process_flow}.

\subsection{Recurrence Memory with CLIP Embeddings}
\label{subsec:recurrence_memory}
Recurrence memory aims to detect when a newly appearing track at time $t$ is a re-appearance of a previously disappeared object. Fig.~\ref{fig:recurrence_flow} illustrates the memory design.

\subsubsection{Active/Dead Memory Buffers}
We maintain:
\begin{itemize}
    \item \textbf{Active memory} $\mathcal{A}$: a dictionary mapping active track IDs to their CLIP embeddings and metadata.
    \item \textbf{Dead memory} $\mathcal{D}$: a FIFO deque of recently disappeared tracks, pruned by a time-to-live (TTL).
\end{itemize}
Each memory entry stores $(\ell,\mathbf{e},t_{\text{last}},t_{\text{anc}})$: label, normalized embedding, last-seen timestamp, and an \emph{anchor frame index} used later for re-conditioning.

\subsubsection{CLIP Crop Embedding}
For a cropped object image $\mathbf{x}$ from frame $\mathbf{I}_t$, we compute:
\begin{equation}
\mathbf{e}(\mathbf{x}) = \frac{f_{\text{CLIP}}(\mathbf{x})}{\|f_{\text{CLIP}}(\mathbf{x})\|_2 + \epsilon},
\end{equation}
\runtfix where $f_{\text{CLIP}}(\cdot)$ is the CLIP image encoder. Embeddings are refreshed periodically (every $\Delta_{\text{upd}}$ frames) to balance cost and robustness.\par\runtoff

\subsubsection{Birth/Death Sets}
Let $\mathcal{I}_t = \{id_i^t\}$ be the set of track IDs at time $t$. We define:
\begin{equation}
\mathcal{B}_t = \mathcal{I}_t \setminus \mathcal{I}_{t-1}, \quad
\mathcal{D}_t = \mathcal{I}_{t-1} \setminus \mathcal{I}_t,
\end{equation}
as the \textbf{birth} and \textbf{death} ID sets, respectively. When $id\in\mathcal{D}_t$, we move its embedding entry from $\mathcal{A}$ to the dead deque $\mathcal{D}$.

\subsubsection{Gated Cosine Similarity Matching}
For each birth $id\in\mathcal{B}_t$, we embed its crop $\mathbf{x}_{id}^t$ and match it against dead candidates of the same label:
\begin{equation}
s(id, j) = \big\langle \mathbf{e}(\mathbf{x}_{id}^t),\ \mathbf{e}_j \big\rangle, \quad
\text{only if } \ell_{id}^t = \ell_j,
\end{equation}
where $\mathbf{e}_j$ is the stored embedding of dead entry $j$. We select $j^\star=\arg\max_j s(id,j)$ and declare recurrence if
\begin{equation}
s(id, j^\star) \ge \sigma,
\end{equation}
\looseness=-1 where $\sigma$ is a similarity threshold. Each successful match increments a recurrence count $R_t$ and retrieves an anchor frame $t_{\text{anc}}^{(j^\star)}$:
\begin{equation}
R_t = \sum_{id\in\mathcal{B}_t} \mathbb{I}\big[s(id, j^\star)\ge \sigma\big],
\quad
\mathcal{A}_t^{\text{anc}} = \{t_{\text{anc}}^{(j^\star)}\}.
\end{equation}
These anchors are used by the boundary manager to augment the Sa2VA batch (Fig.~\ref{fig:recurrence_flow}, ``Segment Builder'').

\subsubsection{TTL Pruning}
\label{subsubsec:ttl_pruning}
Dead memory $\mathcal{D}$ is intended to represent recently disappeared tracks for reliable re-appearance matching; keeping old entries increases false matches (appearance drift) and expands the search set during cosine matching. Thus, we enforce a time-to-live (TTL) window: for a dead entry $j$ with last-seen time $t_{\text{last}}^{(j)}$, we remove it when
\begin{equation}
(t - t_{\text{last}}^{(j)}) > \text{TTL}.
\label{eq:ttl_prune}
\end{equation}
Since $\mathcal{D}$ is stored as a time-ordered deque, pruning is efficient (pop-from-front) and keeps recurrence matching bounded and temporally local.

\looseness=-1 \hl{\textbf{Failure modes and mitigation.} CLIP-based cosine matching can produce false positives under heavy occlusion, significant viewpoint change, or when same-class objects serve as distractors (e.g., multiple persons in a crowded scene), causing the similarity score to be inflated for an incorrect match and potentially triggering a spurious recurrence event. Two mechanisms mitigate this risk. First, \emph{label gating} restricts dead-entry candidates to those sharing the same class label as the new birth, substantially reducing the candidate pool. Second, and most critically, the \emph{identifiability gate} (Sec.}~\ref{subsec:identifiability_gate}\hl{) serves as a second-stage safeguard: if the target is not unambiguously identifiable in the upcoming segment frames, the gate suppresses the refresh call regardless of the recurrence signal, preventing hallucinated masks even when CLIP matching produces a false positive.}

\subsection{Tracking-Driven Event Score}
\label{subsec:event_score}

We compute an event score $E_t$ from tracking state changes and recurrence signals, then apply an EMA smoother to produce a stable trigger signal.

\subsubsection{Birth/Death Counts}
We first compute:
\begin{equation}
b_t = |\mathcal{B}_t|,\quad d_t = |\mathcal{D}_t|.
\end{equation}

\subsubsection{Class Composition Change}
We build a normalized class histogram $\mathbf{h}_t \in \mathbb{R}^C$:
\begin{equation}
h_t(c)=\frac{1}{N_t}\sum_{i=1}^{N_t}\mathbb{I}[\ell_i^t=c],\quad c\in\{1,\dots,C\},
\end{equation}
and define the histogram L1 difference:
\begin{equation}
\Delta H_t = \|\mathbf{h}_t - \mathbf{h}_{t-1}\|_1.
\end{equation}

\subsubsection{Spatial Layout Change via Grid Occupancy}
We discretize the image into a $G\times G$ grid and accumulate a normalized occupancy vector $\mathbf{g}_t \in \mathbb{R}^{G^2}$ by assigning each box center $(c_x,c_y)$ to a grid cell. The layout change is:
\begin{equation}
\Delta L_t = \|\mathbf{g}_t - \mathbf{g}_{t-1}\|_1.
\end{equation}

\subsubsection{Unified Event Score}
Let $R_t$ be the recurrence count from Sec.~\ref{subsec:recurrence_memory}. We define:
\begin{equation}
E_t = w_b b_t + w_d d_t + w_c \Delta H_t + w_l \Delta L_t + w_r R_t,
\label{eq:event_score}
\end{equation}
\looseness=-1 where $\{w_b,w_d,w_c,w_l,w_r\}$ are scalar weights. In our implementation, recurrence is deliberately weighted higher ($w_r$) because a confident re-appearance often warrants immediate refresh.

\subsubsection{EMA Smoothing and Cooldown Trigger}
We smooth $E_t$ using an exponential moving average (EMA):
\begin{equation}
\tilde{E}_t = \alpha E_t + (1-\alpha)\tilde{E}_{t-1},
\label{eq:ema}
\end{equation}
\looseness=-1 \runtfix where $\alpha \in (0,1]$ controls responsiveness. A refresh trigger fires if:
\begin{equation}
\tilde{E}_t \ge \tau \ \wedge\ \texttt{cooldown}=0,
\label{eq:trigger}
\end{equation}
\runtoff where $\tau$ is the event threshold. After triggering, we set \texttt{cooldown}$\leftarrow K$ to suppress immediate re-triggers for $K$ frames. Additionally, we enforce a minimum segment length \texttt{min\_chunk} (Sec.~\ref{subsec:boundary_manager}) to avoid excessively short segments. Recurrence triggers force refresh regardless of $\tilde{E}_t$ (and optionally regardless of cooldown), since re-appearance is a high-risk drift event.

\subsection{Boundary Manager and Segment Construction}
\label{subsec:boundary_manager}

\begin{table*}[t]
\centering
\caption{\textbf{Frozen default event-score weights (kept fixed across all benchmarks).}}
\label{tab:event_weights}
\setlength{\tabcolsep}{7pt}
\renewcommand{\arraystretch}{1.15}
\begin{tabular}{l c p{7.6cm}}
\hline
\textbf{Parameter} & \textbf{Default} & \textbf{Role} \\
\hline
$w_b$ (\texttt{wb}) & 1.0 & Weight for instance births term in the event score. \\
$w_d$ (\texttt{wd}) & 1.0 & Weight for instance deaths term in the event score. \\
$w_c$ (\texttt{wc}) & 0.5 & Weight for class-histogram change term in the event score. \\
$w_l$ (\texttt{wl}) & 0.5 & Weight for spatial layout-grid change term in the event score. \\
$w_r$ (\texttt{wr}) & 2.0 & Weight for recurrence-count term (strong trigger). \\
\hline
\end{tabular}
\end{table*}

The boundary manager converts triggers into segmentation segments. Let accepted boundaries be
\begin{equation}
0=b_0 < b_1 < \cdots < b_K = T,
\end{equation}
\looseness=-1 \runtfix where $b_k$ is accepted only if it satisfies the minimum spacing constraint:
\begin{equation}
b_k - b_{k-1} \ge L_{\min},
\end{equation}
\runtoff with $L_{\min}$ denoting \texttt{min\_chunk}. For each segment $[b_k, b_{k+1})$, we build a Sa2VA input frame index set:
\begin{equation}
\mathcal{J}_k = \Big(\{b_k-p,\dots,b_{k+1}+q\} \cap [0,T-1]\Big)\ \cup\ \mathcal{A}_{b_{k+1}}^{\text{anc}},
\label{eq:batch_idxs}
\end{equation}
where $p,q$ are pre/post context lengths and $\mathcal{A}_{b_{k+1}}^{\text{anc}}$ are anchor frames (possibly empty) obtained from recurrence matching (Fig.~\ref{fig:process_flow}, "Boundary Manager").

\begin{algorithm}[t]
\caption{Event-Driven Sa2VA Inference with Recurrence Anchors}
\label{alg:main}
\begin{algorithmic}[1]
\REQUIRE Video frames $\{\mathbf{I}_t\}_{t=0}^{T-1}$, query $q$, thresholds $\tau,\sigma$, EMA $\alpha$, cooldown $K$, min length $L_{\min}$
\ENSURE Predicted masks $\{\hat{\mathbf{M}}_t\}_{t=0}^{T-1}$
\STATE Decode $(p,\mathcal{V}) \leftarrow \text{LLM}(q)$
\STATE Initialize boundary $b_0\leftarrow 0$, $s\leftarrow 0$ (segment start), $\tilde{E}_{-1}\leftarrow 0$
\FOR{$t=0$ \TO $T-1$}
    \STATE $\mathcal{S}_t \leftarrow \text{Tracker}(\mathbf{I}_t;\mathcal{V})$
    \STATE $(R_t,\mathcal{A}_t^{\text{anc}})\leftarrow \text{RecurrenceUpdate}(t,\mathbf{I}_t,\mathcal{S}_{t-1},\mathcal{S}_t)$
    \STATE Compute $E_t$ via Eq.~\eqref{eq:event_score}; update $\tilde{E}_t$ via Eq.~\eqref{eq:ema}
    \STATE $\text{trig} \leftarrow [\tilde{E}_t\ge \tau\ \wedge\ \texttt{cooldown}=0] \ \vee\ [R_t>0]$
    \IF{$\text{trig} \wedge (t-s)\ge L_{\min}$}
        \STATE Build batch indices $\mathcal{J}$
        \STATE $g\leftarrow \text{Identifiable}(\{\mathbf{I}_u\}_{u\in[s,t)})$
        \IF{$g=\texttt{yes}$}
            \STATE Run Sa2VA on $\{\mathbf{I}_u\}_{u\in\mathcal{J}}$ with prompt $p$ to get $\{\hat{\mathbf{M}}_u\}_{u\in\mathcal{J}}$
        \ELSE
            \STATE Set $\hat{\mathbf{M}}_u\leftarrow \mathbf{0}$ for $u\in\mathcal{J}$
        \ENDIF
        \STATE Re-align and write output masks for core frames $u\in[s,t)$
        \STATE Update boundary: $s\leftarrow t$; set \texttt{cooldown}$\leftarrow K$
    \ENDIF
\ENDFOR
\STATE Process final segment $[s,T)$ similarly
\end{algorithmic}
\end{algorithm}

\begin{algorithm}[t]
\caption{Recurrence Memory Update and Anchor Retrieval}
\label{alg:recurrence}
\begin{algorithmic}[1]
\REQUIRE Time $t$, frame $\mathbf{I}_t$, previous state $\mathcal{S}_{t-1}$, current state $\mathcal{S}_t$, TTL, similarity threshold $\sigma$
\ENSURE Recurrence count $R_t$, anchor set $\mathcal{A}_t^{\text{anc}}$
\STATE Prune dead deque by TTL: remove if $(t-t_{\text{last}})>\text{TTL}$
\STATE Compute births $\mathcal{B}_t$ and deaths $\mathcal{D}_t$ from track ID sets
\STATE Periodically refresh embeddings for active tracks (every $\Delta_{\text{upd}}$ frames)
\FOR{each $id\in\mathcal{D}_t$}
    \STATE Move $(\ell,\mathbf{e},t_{\text{last}},t_{\text{anc}})$ from active dict to dead deque
\ENDFOR
\STATE $R_t\leftarrow 0$, $\mathcal{A}_t^{\text{anc}}\leftarrow \emptyset$
\FOR{each $id\in\mathcal{B}_t$}
    \STATE Embed crop $\mathbf{e}_{id}\leftarrow \text{CLIP}(\text{crop}(\mathbf{I}_t,\mathbf{b}_{id}^t))$
    \STATE Find best dead match $j^\star=\arg\max_{j\in \text{Dead}:\ell_j=\ell_{id}^t}\langle \mathbf{e}_{id},\mathbf{e}_j\rangle$
    \IF{$\langle \mathbf{e}_{id},\mathbf{e}_{j^\star}\rangle \ge \sigma$}
        \STATE $R_t \leftarrow R_t + 1$; $\mathcal{A}_t^{\text{anc}}\leftarrow \mathcal{A}_t^{\text{anc}}\cup\{t_{\text{anc}}^{(j^\star)}\}$
    \ENDIF
    \STATE Register birth as active with its embedding and anchor $t$
\ENDFOR
\RETURN $R_t,\mathcal{A}_t^{\text{anc}}$
\end{algorithmic}
\end{algorithm}

\subsection{Identifiability Gate to Prevent Hallucinated Masks}
\label{subsec:identifiability_gate}
Before running Sa2VA~\cite{sa2va2025}, we perform an identifiability check (InternVL2.5-8B in our implementation). Given a small subset of frames sampled from the \emph{core segment} $[b_k,b_{k+1})$, the gate outputs a binary decision:
\begin{equation}
g_k \in \{\texttt{yes},\texttt{no}\}.
\end{equation}
If $g_k=\texttt{no}$, we return an empty mask sequence for the segment, preventing false positives when the target is absent or not identifiable (Fig.~\ref{fig:process_flow}, ``Identifiability Gate'').

\hl{Concretely, the gate is implemented as a visual question-answering (VQA) prompt to InternVL2.5-8B: given $n_{\text{gate}}{=}3$ uniformly sampled frames from $[b_k, b_{k+1})$ and the original referring query $q$, the model is asked whether the described object is present and unambiguously identifiable. The gate returns $g_k{=}\texttt{yes}$ only upon a positive confirmation; otherwise $g_k{=}\texttt{no}$ and the segment receives empty masks without invoking Sa2VA. This design is intentionally conservative: a missed segment (false negative) is preferable to a hallucinated mask (false positive) when the target is absent or ambiguous. The gate requires no additional threshold tuning beyond the VQA model's own output distribution. The practical impact is quantified in Sec.}~\ref{subsec:gate_analysis}\hl{, where per-dataset rejection rates confirm that the gate actively suppresses 20--26\% of candidate refresh calls across benchmarks.}

\begin{table*}[t]
\centering
\caption{\textbf{EDRRM Configuration Ablation}}
\label{tab:edrrm_ablation}

\setlength{\arrayrulewidth}{1.0pt}

\resizebox{\textwidth}{!}{
\begin{tabular}{l|cccccccccc|ccc}
\hline
\textbf{Config}
& \textbf{thr\_track} & \textbf{ema\_alpha} & \textbf{event\_thr} & \textbf{cooldown} & \textbf{min\_chunk} & \textbf{sim\_thr} & \textbf{ttl}
& \textbf{recurrence\_cooldown} & \textbf{pre\_ctx} & \textbf{post\_ctx}
& \textbf{DAVIS~\cite{davis}} & \textbf{MEVIS~\cite{mevis}} & \textbf{REVOS~\cite{revos}} \\
\hline

edrrm-cfg-1 & 0.5 & 0.3 & 2.5 & 14 & 14 & 0.34 & 110 & 25 & 2 & 2
& 76.78 & 61.50 & 65.34 \\

edrrm-cfg-2 & 0.5 & 0.45 & 2.0 & 10 & 10 & 0.34 & 90 & 25 & 2 & 2
& 76.78 & 61.89 & 65.34 \\

edrrm-cfg-3 & 0.5 & 0.25 & 2.8 & 16 & 16 & 0.36 & 240 & 50 & 2 & 2
& 76.77 & 61.43 & 65.34 \\

edrrm-cfg-4 & 0.42 & 0.45 & 2.0 & 10 & 10 & 0.42 & 110 & 35 & 2 & 2
& 76.76 & 61.08 & 65.33 \\

\palegreen \textbf{edrrm-cfg-5} & \palegreen 0.25 & \palegreen 0.65 & \palegreen 1.4 & \palegreen 4 & \palegreen 4 & \palegreen 0.22 & \palegreen 240 & \palegreen 8 & \palegreen 2 & \palegreen 2
& 76.58 & \palegreen{\textbf{63.13}} & 65.75 \\

edrrm-cfg-6 & 0.3 & 0.7 & 1.6 & 6 & 6 & 0.20 & 300 & 10 & 4 & 4
& 76.21 & 62.43 & 65.60 \\

edrrm-cfg-7 & 0.3 & 0.55 & 1.5 & 4 & 6 & 0.18 & 180 & 6 & 2 & 2
& 76.52 & 62.62 & 65.53 \\

\palegreen \textbf{edrrm-cfg-8} & \palegreen 0.25 & \palegreen 0.45 & \palegreen 1.7 & \palegreen 8 & \palegreen 8 & \palegreen 0.22 & \palegreen 400 & \palegreen 12 & \palegreen 4 & \palegreen 4
& 76.60 & 61.92 & \palegreen{\textbf{66.16}} \\

\palegreen \textbf{edrrm-cfg-9} & \palegreen 0.3 & \palegreen 0.8 & \palegreen 1.2 & \palegreen 2 & \palegreen 4 & \palegreen 0.20 & \palegreen 240 & \palegreen 4 & \palegreen 8 & \palegreen 8
& \palegreen{\textbf{76.84}} & 60.92 & 65.55 \\

\hline
\end{tabular}
}
\end{table*}

\subsection{Sa2VA Mask Prediction and Output Alignment}
\label{subsec:sa2va}
If the segment passes the gate, we invoke Sa2VA on the batch frames $\{\mathbf{I}_t\}_{t\in\mathcal{J}_k}$ with prompt $p$:
\begin{equation}
\{\hat{\mathbf{M}}_t\}_{t\in\mathcal{J}_k} = f_{\text{Sa2VA}}\big(\{\mathbf{I}_t\}_{t\in\mathcal{J}_k},\ p\big).
\end{equation}
\looseness=-1 Since $\mathcal{J}_k$ may include anchors and context frames, we \textbf{re-align} the predicted masks to the contiguous core indices $t\in[b_k,b_{k+1})$ by indexing into the batch result (implementation detail in our runner). The global output is the concatenation over all segments:
\begin{equation}
\hat{\mathbf{M}}_t = \hat{\mathbf{M}}^{(k)}_t,\quad \forall t\in[b_k,b_{k+1}).
\end{equation}

\subsection{Algorithms}
\label{subsec:algorithms}
Algorithm~\ref{alg:main} summarizes the full event-driven inference loop (Fig.~\ref{fig:process_flow}): for each frame, we update tracking, compute the event score and EMA, and accept a boundary only when the trigger fires and the minimum segment length constraint holds; the boundary manager then builds the Sa2VA batch indices (core segment plus optional context and recurrence anchors) and runs an identifiability gate before invoking Sa2VA, finally re-aligning batch masks to the core timeline. Algorithm~\ref{alg:recurrence} details recurrence memory (Fig.~\ref{fig:recurrence_flow}): deaths move from active to dead, dead entries are TTL-pruned (Eq.~\ref{eq:ttl_prune}), and each birth is embedded with CLIP and matched (label-gated cosine) against dead candidates; successful matches yield a recurrence count and anchor frames that are injected into the next segment batch.

\subsection{Parameters and Defaults}
\label{subsec:hyperparams}
Table~\ref{tab:event_weights} reports the parameter defaults that we keep frozen across all dataset benchmarks: the event-score weighting parameters $(w_b,w_d,w_c,w_l,w_r)$.
These weights define the relative contribution of instance births, instance deaths, class-composition change, layout change, and recurrence signals in our event score, ensuring a consistent definition of ``eventfulness'' across Ref-DAVIS17~\cite{davis}, MeViS~\cite{mevis}, and ReVOS~\cite{revos}.

\subsection{Computational Cost}
\label{subsec:complexity}

At time $t$, event score computation is $O(N_t + G^2 + C)$ for births/deaths, grid occupancy, and histogram updates. Recurrence matching cost is dominated by comparing each birth embedding to dead candidates; with label gating, the worst-case cost is $O(|\mathcal{B}_t|\cdot|\mathcal{D}|)$ but typically much smaller due to TTL pruning and label filtering. The overall system reduces expensive Sa2VA calls by segmenting only selected batches, yielding an efficiency-accuracy trade-off governed by $\tau,\alpha,K,L_{\min},\sigma$, and TTL.

\section{Experiments}
\label{sec:guidelines}
We evaluate Event-Driven Refresh + Recurrence Memory (EDRRM) on Ref-DAVIS17~\cite{davis}, MeViS~\cite{mevis}, and ReVOS~\cite{revos} using J\&F as the primary metric, and compare against two refresh schedulers: Fixed Window and FrameDiff-SSIM~\cite{ssim2004}. All experiments are run on a machine with 4$\times$ NVIDIA RTX 8000 GPUs. We first visualize how event refresh selects boundaries and how each score component contributes (Sec.~\ref{subsec:event_timeline}), then report configuration ablations (Sec.~\ref{subsec:ablation_baselines}). Next, we analyze the accuracy-cost trade-off via the average number of Sa2VA calls~\cite{sa2va2025} and Pareto optimality (Sec.~\ref{subsec:pareto}), followed by refresh-rate budget evaluation (Sec.~\ref{subsec:budget}). Finally, we present main results under global-tuned vs.\ per-dataset oracle tuning and provide a qualitative comparison.

\begin{figure*}[t]
  \centering
  \includegraphics[width=0.945\textwidth]{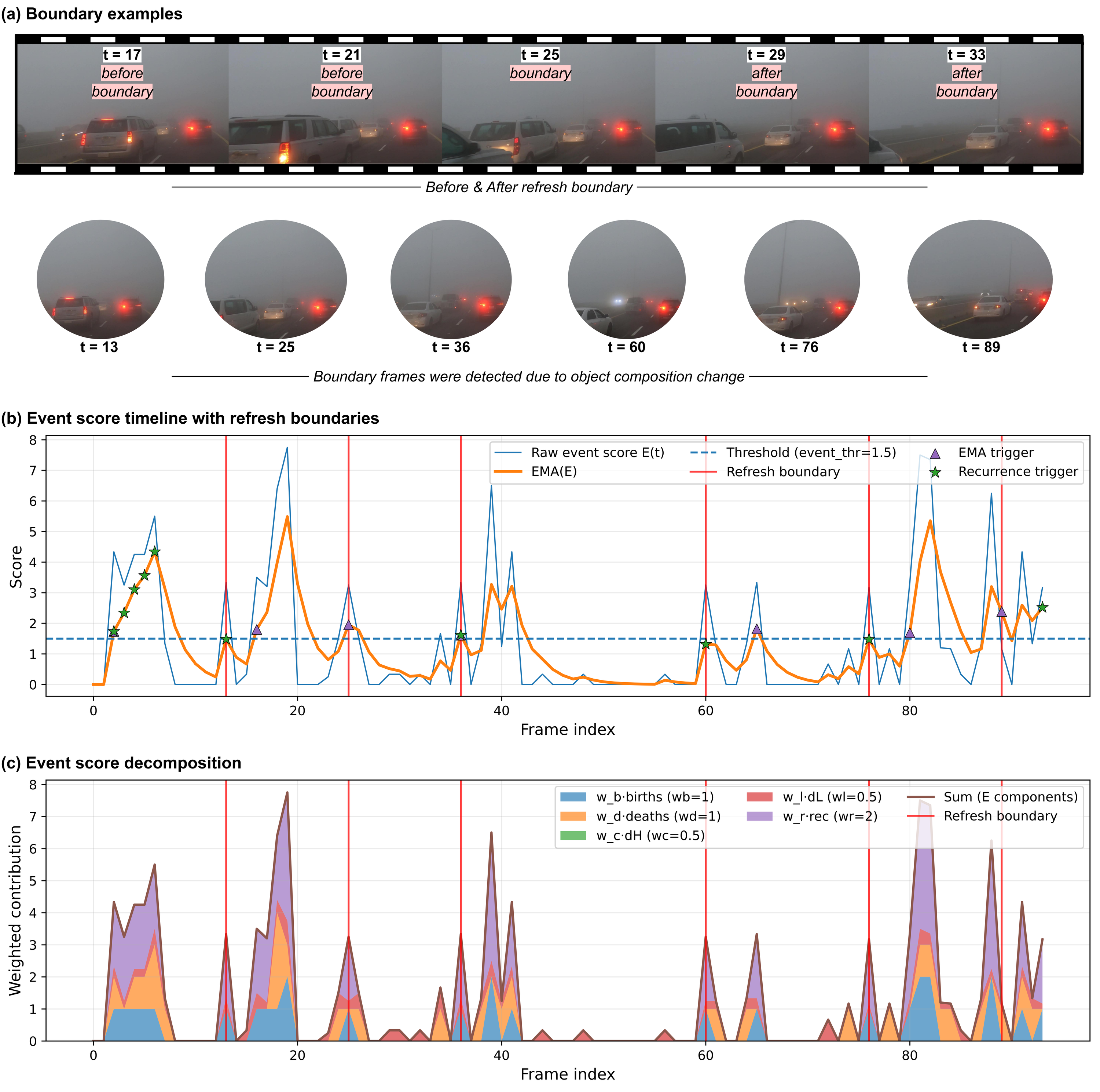}
  \caption{\textbf{Event-refresh visualization and interpretability.}}
  \label{fig:event_timeline}
\end{figure*}

\begin{table}[H]
\centering
\caption{\textbf{FrameDiff-SSIM Configuration Ablation}}
\label{tab:ssim_ablation}
\setlength{\arrayrulewidth}{1.0pt}

\resizebox{\columnwidth}{!}{
\begin{tabular}{l|ccc|ccc}
\hline
\textbf{Config} & \textbf{threshold} & \textbf{cooldown} & \textbf{min\_length}
& \textbf{DAVIS~\cite{davis}} & \textbf{MEVIS~\cite{mevis}} & \textbf{REVOS~\cite{revos}} \\
\hline

\palegreen \textbf{ssim-cfg-1} & \palegreen 0.95 & \palegreen 4 & \palegreen 4
& 76.83 & \palegreen{\textbf{62.48}} & 65.25 \\

\palegreen \textbf{ssim-cfg-2} & \palegreen 0.93 & \palegreen 8 & \palegreen 8
& \palegreen{\textbf{77.39}} & 62.24 & \palegreen{\textbf{65.55}} \\

ssim-cfg-3 & 0.90 & 8 & 12
& 76.85 & 61.76 & 65.41 \\

ssim-cfg-4 & 0.87 & 12 & 16
& 76.79 & 61.59 & 65.52 \\

ssim-cfg-5 & 0.82 & 20 & 32
& 76.47 & 59.85 & 65.43 \\

ssim-cfg-6 & 0.80 & 8 & 8
& 77.24 & 62.35 & 65.21 \\

ssim-cfg-7 & 0.30 & 8 & 8
& 76.48 & 59.91 & 65.22 \\

\hline
\end{tabular}
}
\end{table}

\begin{table}[H]

\centering
\caption{\textbf{Fixed Window Configuration Ablation}}
\label{tab:window_ablation}
\setlength{\arrayrulewidth}{1.0pt}

\resizebox{0.78\columnwidth}{!}{
\begin{tabular}{l|c|ccc}
\hline
\textbf{Config} & \textbf{length} & \textbf{DAVIS~\cite{davis}} & \textbf{MEVIS~\cite{mevis}} & \textbf{REVOS~\cite{revos}} \\
\hline

\palegreen \textbf{window-cfg-1} & \palegreen 8
& \palegreen{\textbf{77.45}} & \palegreen{\textbf{62.34}} & 65.55 \\

\palegreen \textbf{window-cfg-2} & \palegreen 16
& 76.53 & 55.25 & \palegreen{\textbf{65.88}} \\

window-cfg-3 & 32
& 76.99 & 59.85 & 65.40 \\

window-cfg-4 & 64
& 75.94 & 58.92 & 65.02 \\

window-cfg-5 & 128
& 76.31 & 59.00 & 65.40 \\

\hline
\end{tabular}
}
\end{table}

\subsection{\hl{Implementation Details}}
\label{subsec:impl_details}
\looseness=-1 \hl{All experiments are run on 4$\times$ NVIDIA RTX 8000 GPUs. The object tracker is MASA}~\cite{li2024masa}\hl{ with an open-vocabulary detector backbone; \texttt{thr\_track} controls the minimum track confidence and is swept from 0.25 to 0.50 in our sensitivity ablation (Sec.}~\ref{subsec:tracker_sensitivity}\hl{). CLIP embeddings for recurrence matching use the ViT-B/32 encoder}~\cite{radford2021clip}\hl{ with object crops resized to $224{\times}224$. The identifiability gate uses InternVL2.5-8B with $n_{\text{gate}}{=}3$ sampled frames per segment. Query decoding uses InternLM2.5-7B (packaged within Sa2VA-8B), so EDRRM requires no additional large models beyond Sa2VA-8B. The default global-tuned configuration (\texttt{edrrm-cfg-5}) uses: $\texttt{thr\_track}{=}0.25$, $\alpha{=}0.65$, $\tau{=}1.4$, $\texttt{cooldown}{=}4$, $\texttt{min\_chunk}{=}4$, $\texttt{sim\_thr}{=}0.22$, $\texttt{TTL}{=}240$, $\texttt{recurrence\_cooldown}{=}8$, $\texttt{pre\_ctx}{=}2$, and $\texttt{post\_ctx}{=}2$. All hyperparameters were tuned by grid search over the MeViS validation split and then frozen for DAVIS and ReVOS under the global-tuned protocol, or independently selected per dataset under the oracle-tuned protocol. No dataset-specific feature engineering beyond threshold tuning was applied.}

\subsection{Event Refresh Timeline and Score Decomposition}
\label{subsec:event_timeline}
Fig.~\ref{fig:event_timeline} illustrates how our event-driven scheduler produces refresh boundaries over time and why each boundary is selected. In Fig.~\ref{fig:event_timeline}(a), frames surrounding a boundary show a clear and persistent change in the scene/object composition, which motivates re-grounding rather than continuing long-range propagation. Fig.~\ref{fig:event_timeline}(b) plots the raw event score $E(t)$ together with its EMA-smoothed signal; a boundary is accepted when the EMA exceeds the threshold and the minimum segment-length constraint is satisfied, while recurrence triggers are marked when a previously disappeared instance is matched and yields anchor frames. Finally, Fig.~\ref{fig:event_timeline}(c) decomposes $E(t)$ into its weighted components (births, deaths, class/layout change, and recurrence), showing that boundary locations align with large contributions from one or more cues, thus providing interpretability for the refresh decisions.

\subsection{Configuration Ablation and Baseline Comparison}
\label{subsec:ablation_baselines}

\looseness=-1 Tables~\ref{tab:edrrm_ablation},~\ref{tab:ssim_ablation}, and~\ref{tab:window_ablation} summarize our configuration sweeps for the proposed Event-Driven Refresh + Recurrence Memory (EDRRM) and two competing scheduling frameworks: FrameDiff-SSIM~\cite{ssim2004} and Fixed Window. For EDRRM (Table~\ref{tab:edrrm_ablation}), we jointly vary the tracker filtering and event scheduler (\texttt{thr\_track}, $\alpha$, $\tau$, \texttt{cooldown}, \texttt{min\_chunk}) together with recurrence controls (\texttt{sim\_thr}, \texttt{ttl}, \texttt{recurrence\_cooldown}) and optional context (\texttt{pre\_ctx}/\texttt{post\_ctx}); the best per-dataset performance is achieved by \textbf{edrrm-cfg-9} on Ref-DAVIS17 (\textbf{76.84}), \textbf{edrrm-cfg-5} on MeViS (\textbf{63.13}), and \textbf{edrrm-cfg-8} on ReVOS (\textbf{66.16}). For FrameDiff-SSIM (Table~\ref{tab:ssim_ablation}), we sweep the SSIM threshold and temporal constraints, obtaining best Ref-DAVIS17 and ReVOS with \textbf{ssim-cfg-2} (\textbf{77.39}, \textbf{65.55}) and best MeViS with \textbf{ssim-cfg-1} (\textbf{62.48}). For Fixed Window (Table~\ref{tab:window_ablation}), the window length controls refresh frequency, where \textbf{window-cfg-1} gives the best Ref-DAVIS17/MeViS (\textbf{77.45}, \textbf{62.34}) and \textbf{window-cfg-2} gives the best ReVOS (\textbf{65.88}). Overall, EDRRM achieves the strongest results on MeViS and ReVOS among the evaluated frameworks, while remaining competitive on Ref-DAVIS17, highlighting the benefit of refreshing at content-driven change points.

\subsection{\hl{Component Ablation}}
\label{subsec:component_ablation}
\begin{figure*}[t]
  \centering
  \includegraphics[width=\textwidth]{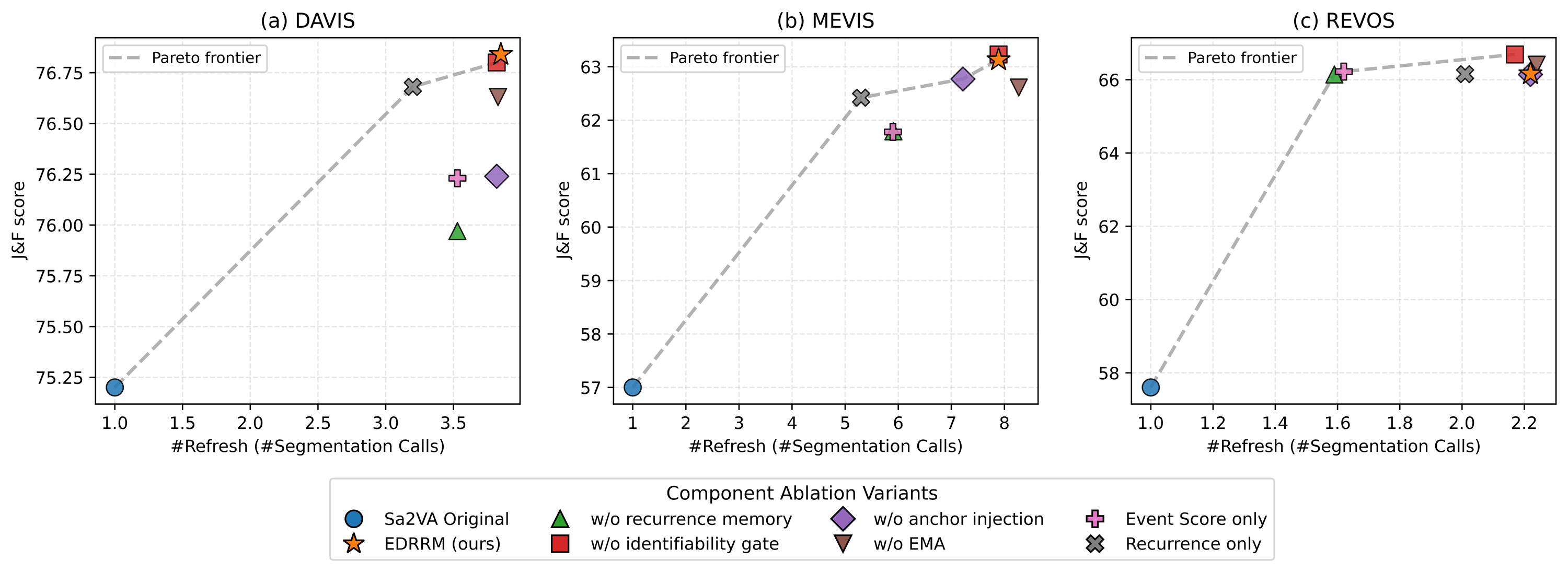}
  \caption{\hl{\textbf{Component ablation Pareto frontier.} J\&F versus average \#refresh for all ablation variants on Ref-DAVIS17 (a), MeViS (b), and ReVOS (c). The dashed curve marks the Pareto frontier. EDRRM (Ours) consistently lies on or near the frontier, confirming that the full system achieves the best accuracy-efficiency balance among all variants.}}
  \label{fig:component_ablation_pareto}
\end{figure*}

\begin{figure*}[t]
  \centering
  \includegraphics[width=\textwidth]{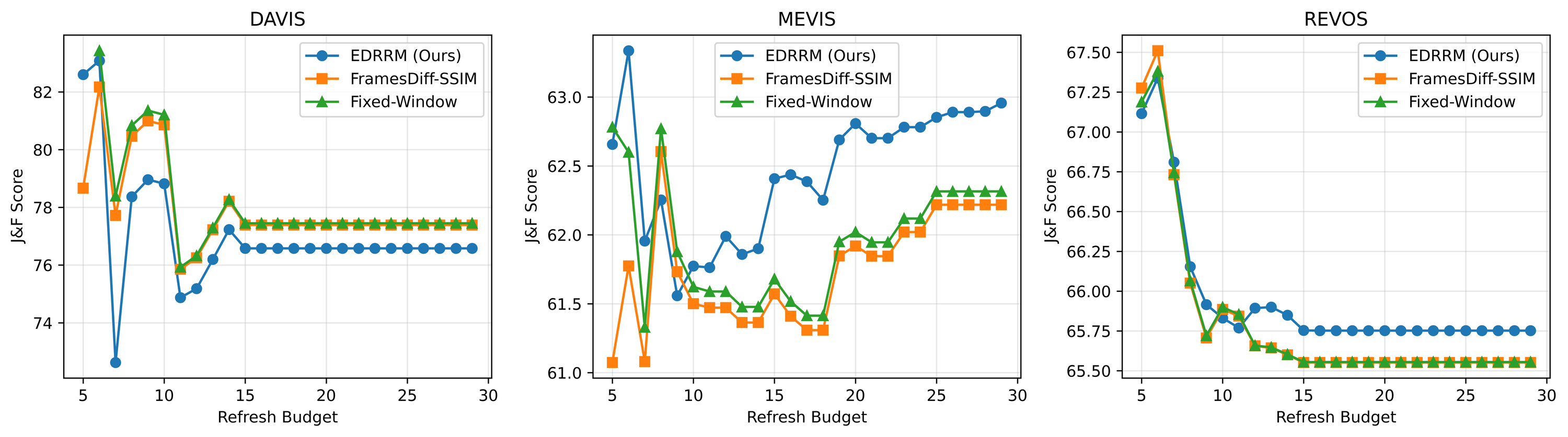}
  \caption{\hl{\textbf{J\&F score versus integer refresh budget (5--30 Sa2VA calls).} EDRRM, FrameDiff-SSIM, and Fixed-Window on Ref-DAVIS17, MeViS, and ReVOS. EDRRM maintains higher J\&F across a broad budget range, particularly in MeViS, demonstrating that event-driven scheduling outperforms fixed-schedule and appearance-based baselines at matched inference cost.}}
  \label{fig:refresh_budget_curves}
\end{figure*}

\hl{To isolate the contribution of each module, Table}~\ref{tab:component_ablation}\hl{ reports J\&F and average \#refresh (\#ref) for eight variants on all three benchmarks. The Pareto frontier in Fig.}~\ref{fig:component_ablation_pareto}\hl{ visualizes accuracy versus refresh cost per variant, and Fig.}~\ref{fig:refresh_budget_curves}\hl{ shows J\&F as a function of integer refresh budget across the three methods.}

\begin{table*}[t]
\centering
\caption{\hl{\textbf{Component Ablation.} J\&F (\%) and average \#refresh (\#ref) for each variant on Ref-DAVIS17, MeViS, and ReVOS under the oracle-tuned best configuration. Each ablated row removes exactly one module from the full EDRRM system.}}
\label{tab:component_ablation}
\setlength{\arrayrulewidth}{1.0pt}
\setlength{\tabcolsep}{6pt}
\renewcommand{\arraystretch}{1.15}
\small
\begin{tabular}{l|cc|cc|cc}
\hline
\multirow{2}{*}{\textbf{Variant}} &
\multicolumn{2}{c|}{\textbf{DAVIS~\cite{davis}}} &
\multicolumn{2}{c|}{\textbf{MeViS~\cite{mevis}}} &
\multicolumn{2}{c}{\textbf{ReVOS~\cite{revos}}} \\
& \textbf{J\&F} & \textbf{\#ref} & \textbf{J\&F} & \textbf{\#ref} & \textbf{J\&F} & \textbf{\#ref} \\
\hline
Sa2VA Original           & 75.20 & 1.00 & 57.00 & 1.00 & 57.60 & 1.00 \\
\hline
\textbf{EDRRM (Ours)}    & \textbf{76.84} & 3.85 & \textbf{63.13} & 7.89 & \textbf{66.16} & 2.22 \\
w/o recurrence memory    & 75.97 & 3.53 & 61.79 & 5.91 & 66.14 & 1.59 \\
w/o identifiability gate & 76.80 & 3.82 & 63.23 & 7.89 & 66.69 & 2.17 \\
w/o anchor injection     & 76.24 & 3.82 & 62.77 & 7.22 & 66.14 & 2.22 \\
w/o EMA                  & 76.63 & 3.83 & 62.61 & 8.27 & 66.41 & 2.24 \\
Event Score only         & 76.23 & 3.53 & 61.78 & 5.90 & 66.22 & 1.62 \\
Recurrence only          & 76.68 & 3.20 & 62.42 & 5.30 & 66.15 & 2.01 \\
\hline
\end{tabular}
\end{table*}

\looseness=-1 \hl{Several patterns emerge from Table}~\ref{tab:component_ablation}\hl{. First, comparing \emph{Event Score only} against \emph{Recurrence only} directly separates the two core contributions: Event Score only achieves 76.23/61.78/66.22 while Recurrence only achieves 76.68/62.42/66.15, and the full EDRRM combining both reaches 76.84/63.13/66.16, confirming that gains arise from both general event-driven scheduling \emph{and} specific re-appearance handling, not from either alone. Second, removing recurrence memory reduces MeViS J\&F from 63.13 to 61.79 while also reducing refresh cost (7.89$\rightarrow$5.91 \#ref), indicating that recurrence events trigger useful additional refreshes that recover from re-appearance failures. Third, the identifiability gate's contribution is primarily in false-positive suppression (discussed further in Sec.}~\ref{subsec:gate_analysis}\hl{) rather than bulk J\&F, as evidenced by the modest raw J\&F difference when removing it.}

\subsection{\hl{Tracker Sensitivity Analysis}}
\label{subsec:tracker_sensitivity}

\begin{figure*}[t]
  \centering
  \includegraphics[width=\textwidth]{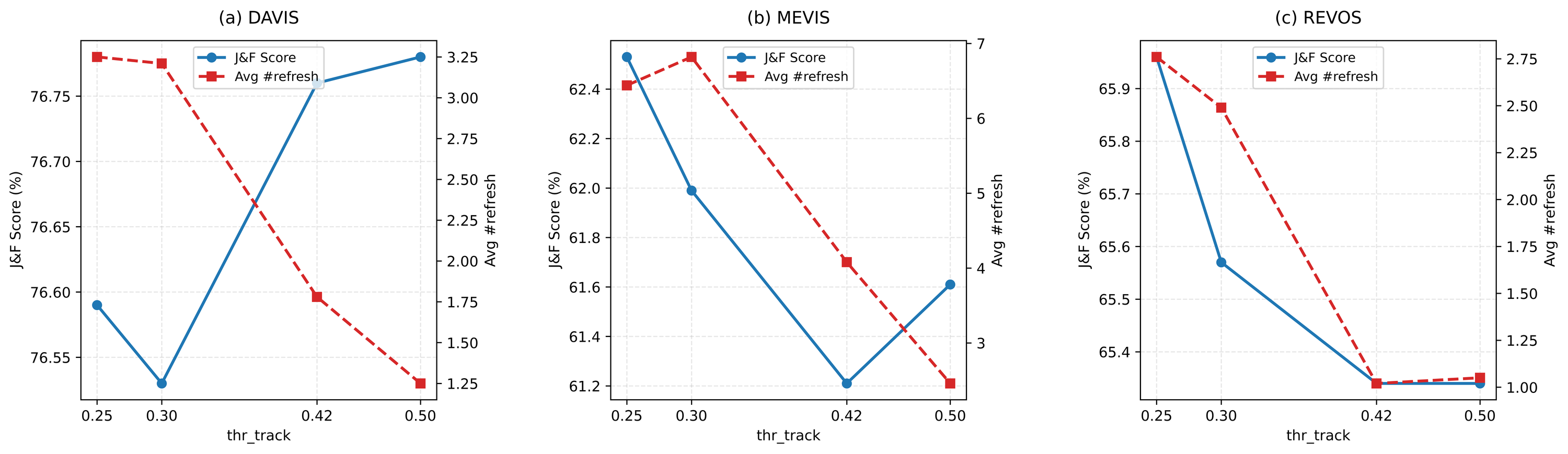}
  \caption{\hl{\textbf{Tracker sensitivity analysis.} J\&F score (blue, left axis) and average \#refresh (red dashed, right axis) versus \texttt{thr\_track} on Ref-DAVIS17 (a), MeViS (b), and ReVOS (c). Higher \texttt{thr\_track} reduces refresh count but also suppresses valid events, creating a graceful accuracy-cost trade-off.}}
  \label{fig:thr_track_sensitivity}
\end{figure*}

\hl{Since all event signals derive from tracker output, we analyze how tracking strictness affects EDRRM. Table}~\ref{tab:tracker_sensitivity}\hl{ sweeps \texttt{thr\_track} from 0.25 (Very noisy: many tracks accepted) to 0.50 (Very Strict: only high-confidence tracks), effectively modeling tracker reliability. Fig.}~\ref{fig:thr_track_sensitivity}\hl{ shows J\&F and average \#refresh jointly as \texttt{thr\_track} increases.}

\begin{table*}[t]
\centering
\caption{\hl{\textbf{Tracker Sensitivity Analysis.} Effect of \texttt{thr\_track} on J\&F (\%) and average \#refresh (\#ref). Higher \texttt{thr\_track} produces stricter, less noisy tracking with fewer but more reliable events.}}
\label{tab:tracker_sensitivity}
\setlength{\arrayrulewidth}{1.0pt}
\setlength{\tabcolsep}{6pt}
\renewcommand{\arraystretch}{1.15}
\small
\begin{tabular}{c|l|cc|cc|cc}
\hline
\multirow{2}{*}{\textbf{thr\_track}} & \multirow{2}{*}{\textbf{Tracking Quality}} &
\multicolumn{2}{c|}{\textbf{DAVIS~\cite{davis}}} &
\multicolumn{2}{c|}{\textbf{MeViS~\cite{mevis}}} &
\multicolumn{2}{c}{\textbf{ReVOS~\cite{revos}}} \\
& & \textbf{J\&F} & \textbf{\#ref} & \textbf{J\&F} & \textbf{\#ref} & \textbf{J\&F} & \textbf{\#ref} \\
\hline
0.25 & Very noisy  & 76.59 & 3.25 & 62.53 & 6.44 & 65.96 & 2.76 \\
0.30 & Noisy       & 76.53 & 3.21 & 61.99 & 6.82 & 65.57 & 2.49 \\
0.42 & Strict      & 76.76 & 1.78 & 61.21 & 4.08 & 65.34 & 1.02 \\
0.50 & Very Strict & 76.78 & 1.25 & 61.61 & 2.46 & 65.34 & 1.05 \\
\hline
\end{tabular}
\end{table*}

\hl{At low \texttt{thr\_track} (Very noisy), more tracks generate more events and higher refresh counts, recovering more re-appearance events and yielding higher J\&F on MeViS (62.53) and ReVOS (65.96). At high \texttt{thr\_track} (Very Strict), fewer tracks survive, reducing refresh calls dramatically (to 1.25 on DAVIS) but suppressing valid events. Critically, J\&F degrades \emph{gracefully}: the accuracy range across all \texttt{thr\_track} values is at most 1.3 J\&F points on any dataset, demonstrating that EDRRM is not brittle to this parameter. This proxy analysis models the effect of tracker reliability (e.g., missed detections or ID switches reducing track completeness): even under degraded tracking, accuracy remains within a tight and predictable band. While direct injection of ID-switch or missed-detection failures is left for future work, the \texttt{thr\_track} sweep captures the accuracy-robustness trade-off under degraded track completeness as a tractable proxy.}

\subsection{\hl{Identifiability Gate Analysis}}
\label{subsec:gate_analysis}

\hl{Table}~\ref{tab:gate_statistics}\hl{ reports the total candidate refresh segments, the number suppressed by the gate ($g_k{=}\texttt{no}$), and the gate rejection rate per benchmark.}

\begin{table}[t]
\centering
\caption{\hl{\textbf{Identifiability Gate Statistics.} Candidate refresh segments suppressed by the gate under the default EDRRM configuration (\texttt{edrrm-cfg-5}).}}
\label{tab:gate_statistics}
\setlength{\arrayrulewidth}{1.0pt}
\setlength{\tabcolsep}{8pt}
\renewcommand{\arraystretch}{1.15}
\small
\begin{tabular}{l|c|c|c}
\hline
\textbf{Dataset} & \textbf{Total Segments} & \textbf{Gate ``no''} & \textbf{Gate ``no'' Rate} \\
\hline
DAVIS & 934  & 192  & 20.56\% \\
MeViS & 6224 & 1383 & 22.22\% \\
ReVOS & 6691 & 1713 & 25.60\% \\
\hline
\end{tabular}
\end{table}

\looseness=-1 \hl{The gate suppresses 20--26\% of candidate refresh calls per dataset, confirming that a substantial fraction of triggered events correspond to segments where the target is absent or unidentifiable. Without the gate, these segments would invoke Sa2VA unnecessarily, generating hallucinated masks. The higher rejection rate on ReVOS (25.60\%) versus DAVIS (20.56\%) is consistent with ReVOS being a more dynamic benchmark with more frequent target absence. A gate false reject suppresses a segment where the target is actually present and reduces J\&F by forcing empty masks; the net J\&F effect is therefore determined by the balance between false-positive prevention and false-reject cost, which explains the modest raw J\&F difference in Table}~\ref{tab:component_ablation}.

\subsection{\hl{Refresh Trigger Analysis}}
\label{subsec:trigger_analysis}

\begin{figure*}[t]
  \centering
  \includegraphics[width=\textwidth]{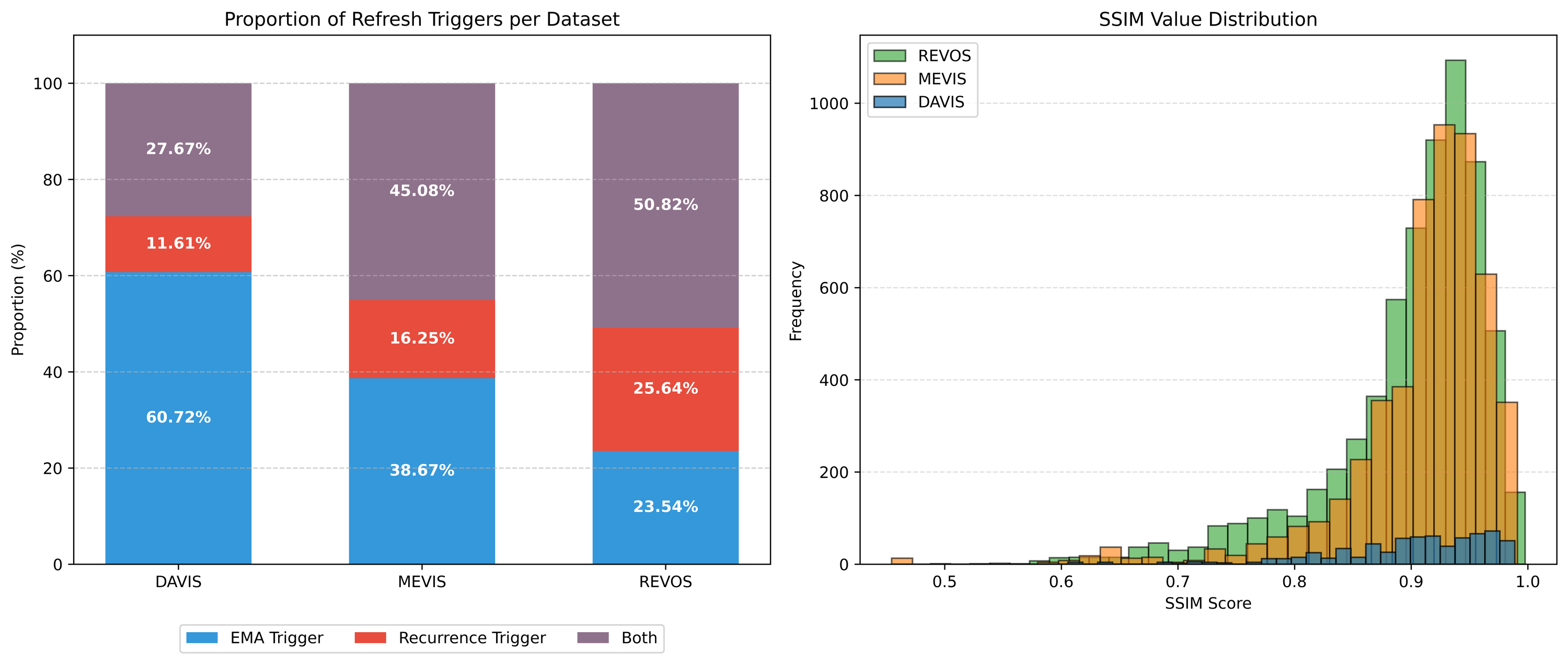}
  \caption{\hl{\textbf{Refresh trigger analysis.} \textit{Left}: Proportion of refresh triggers attributed to EMA event score only (blue), recurrence matching only (red), or both simultaneously (purple) per dataset. \textit{Right}: SSIM value distribution across all consecutive frame pairs in each benchmark, illustrating why SSIM-based detection is threshold-sensitive.}}
  \label{fig:trigger_analysis}
\end{figure*}

\hl{Fig.}~\ref{fig:trigger_analysis}\hl{ (left) decomposes refresh triggers into: pure EMA-triggered (event score), pure recurrence-triggered, and frames where both fire simultaneously. On DAVIS, 60.72\% of triggers stem from the EMA event score alone, reflecting shorter videos with fewer re-appearance events. On MeViS and ReVOS, ``Both'' triggers grow to 45.08\% and 50.82\% respectively, consistent with longer and more dynamic videos. This decomposition directly answers the question of whether improvement comes mainly from the general event score or from re-appearance handling: both contribute, with recurrence handling becoming increasingly important in longer, more complex benchmarks, exactly matching the J\&F patterns in Table}~\ref{tab:component_ablation}\hl{. The SSIM distribution (right) further shows why SSIM-based baselines are threshold-sensitive: SSIM values cluster near 1.0 for all datasets, leaving a very narrow effective detection range.}

\begin{figure*}[t]
  \centering
  \includegraphics[width=0.97\textwidth]{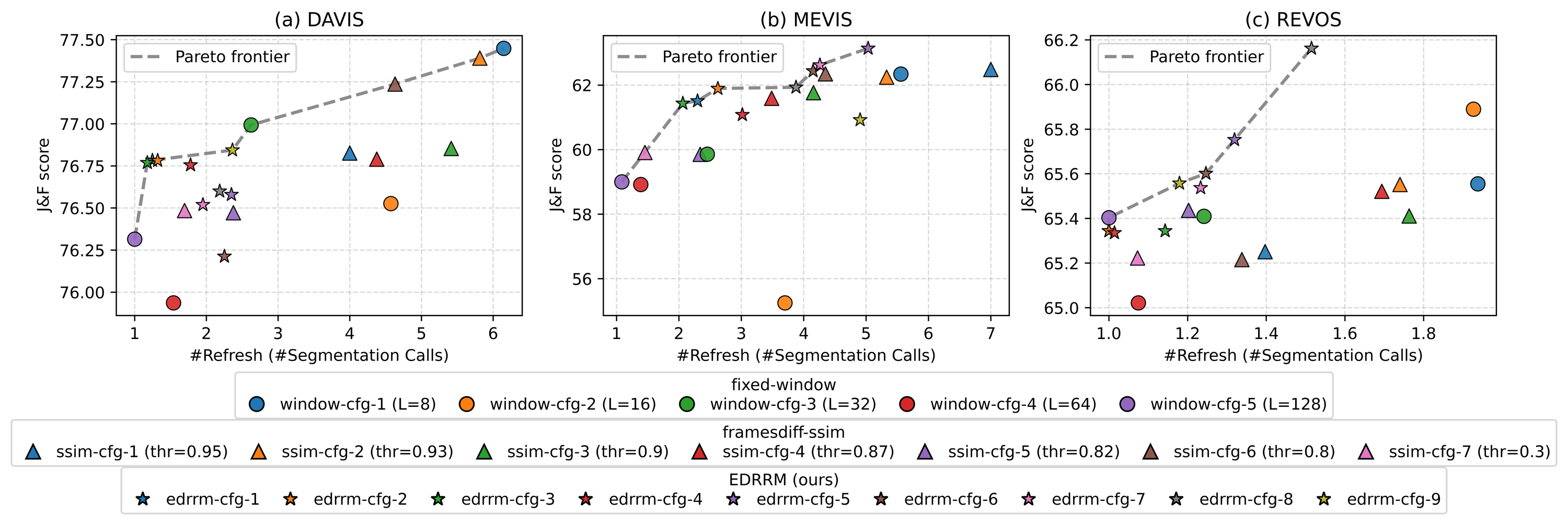}
  \caption{\textbf{J\&F score versus average \#refresh (Sa2VA segmentation calls) for all configurations.}}
  \label{fig:pareto_trade_off}
\end{figure*}

\begin{figure*}[t]
  \centering
  \includegraphics[width=\textwidth]{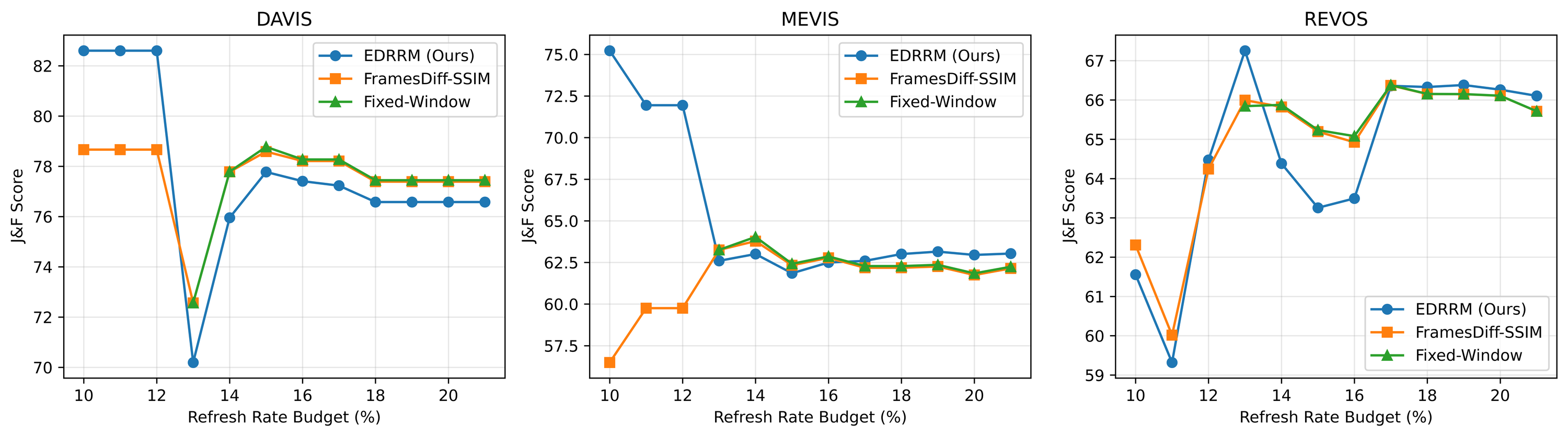}
  \caption{\textbf{Budget-conditioned mean J\&F versus refresh-rate budget.} Note: Budgets where this shared intersection is empty yield undefined means; we therefore report the non-empty range (10--22\%) in our benchmarks.}

  \label{fig:refresh_rate_plot}
\end{figure*}

\subsection{Pareto Trade-off Between Accuracy and Refresh Cost}
\label{subsec:pareto}

\looseness=-1 Fig.~\ref{fig:pareto_trade_off} analyzes the accuracy-cost trade-off by plotting each configuration as a point in the plane of average \#refresh (i.e., the average number of Sa2VA segmentation calls per video) versus J\&F score. Since each refresh triggers an additional grounded segmentation inference, fewer refreshes directly translate to lower compute and latency. The dashed curve denotes the Pareto frontier, where no configuration can be improved in J\&F without increasing refresh cost. Notably, many of our EDRRM configurations lie on or close to this frontier across Ref-DAVIS17~\cite{davis}, MeViS~\cite{mevis}, and ReVOS~\cite{revos}, indicating that adaptive, event-driven boundaries achieve near-optimal accuracy for a given number of segmentation calls. This motivates our design choice: rather than refreshing at fixed intervals (Fixed Window) or relying on low-level appearance change heuristics (FrameDiff-SSIM), detecting \emph{object-composition change} enables a segment sampler that concentrates Sa2VA calls at semantically meaningful moments, yielding a better accuracy-efficiency balance.

\begin{table*}[t]
\centering
\caption{\textbf{Comparison of refresh rate budget, segment length statistics, and J\&F scores.}}
\label{tab:budget_len_jf_summary}

\setlength{\arrayrulewidth}{1.0pt}
\setlength{\tabcolsep}{5pt}
\renewcommand{\arraystretch}{1.15}

\begin{tabular}{c|ccc|ccc|ccc|ccc}
\hline
\multirow{2}{*}{\textbf{Method}} &
\multicolumn{3}{c|}{\textbf{Refresh Rate Budget $\downarrow$ }} &
\multicolumn{3}{c|}{\textbf{Segment Length (Avg) $\uparrow$ }} &
\multicolumn{3}{c|}{\textbf{Segment Length (Min/Max) $\uparrow$ }} &
\multicolumn{3}{c}{\textbf{Peak budget-conditioned J\&F $\uparrow$ }} \\
& \textbf{DAVIS} & \textbf{MEVIS} & \textbf{REVOS} &
  \textbf{DAVIS} & \textbf{MEVIS} & \textbf{REVOS} &
  \textbf{DAVIS} & \textbf{MEVIS} & \textbf{REVOS} &
  \textbf{DAVIS} & \textbf{MEVIS} & \textbf{REVOS} \\
\hline

Window &
15 & 14 & 17 &
7.58 & 7.52 & 6.72 &
4/8 & 4/8 & 3/7 &
78.78 & 64.03 & 66.38 \\
\hline

SSIM &
10 & 14 & 17 &
8.04 & 7.83 & 7.85 &
4/9 & 5/8 & 4/9 &
78.66 & 63.78 & 66.37 \\
\hline

\textbf{Ours} &
\textbf{10} & \textbf{10} & \textbf{13} &
\textbf{36.37} & \textbf{9.97} & \textbf{13.79} &
\textbf{30/44} & \textbf{5/18} & \textbf{11/16} &
\textbf{82.60} & \textbf{75.22} & \textbf{67.25} \\
\hline

\end{tabular}
\end{table*}

\subsection{Refresh Rate Budget and Budget-Conditioned J\&F Efficiency}
\label{subsec:budget}

Fig.~\ref{fig:refresh_rate_plot} and Table~\ref{tab:budget_len_jf_summary} summarize efficiency under a refresh-rate budget using \emph{fixed} tuned settings. We instantiate each framework with a single best-tuned configuration from the ablations in Sec.~\ref{subsec:ablation_baselines}: Fixed Window uses length $L{=}8$ (\texttt{window-cfg-1}), FrameDiff-SSIM uses threshold $0.93$ (\texttt{ssim-cfg-2}), and EDRRM uses $\texttt{thr\_track}{=}0.25$ (\texttt{edrrm-cfg-5}). For a video instance $v$, we define its refresh rate as $r(v)=100\cdot N_{\text{refresh}}(v)/T(v)$, where $N_{\text{refresh}}(v)$ is the number of refreshes (Sa2VA segmentation calls) and $T(v)$ is the number of frames.

Given a budget $B\in[0,100]$, we form (for each method) the subset of video instances whose refresh rate satisfies $r(v)<B$. To ensure a fair comparison at each budget, we evaluate all methods on the \emph{intersection} of these subsets (i.e., the same set of instances shared across methods at that budget). The plotted value in Fig.~\ref{fig:refresh_rate_plot} is then the \emph{budget-conditioned} mean J\&F over this common subset. These budget-conditioned values are computed on a budget-filtered shared subset and are therefore not directly comparable to full-dataset averages in Tables~2--4 and~6--7. Table~\ref{tab:budget_len_jf_summary} reports, for each dataset and method, the budget $B^\star$ at which this budget-conditioned mean J\&F is maximized, along with the corresponding peak J\&F value. We also report segment-length statistics for the same fixed configurations; segment-length statistics are computed over the full benchmark under the same fixed configuration (not budget-filtered). Longer segment lengths are desirable since they imply fewer boundary splits (fewer segmentation calls) and more stable within-segment propagation.

Overall, our method reaches its peak J\&F at smaller budgets than both baselines, meaning that high accuracy is achieved with fewer Sa2VA calls. This is practically important because a smaller refresh budget reduces the number of segmentation calls, lowering runtime and compute while maintaining strong RVOS accuracy.

\looseness=-1 \hl{Table}~\ref{tab:matched_budget}\hl{ further provides an apples-to-apples matched-budget comparison by grouping video instances into budget bins by average \#calls (${\leq}2$, ${\leq}4$, ${\leq}8$) and reporting J\&F only for videos within each bin, evaluated on the intersection of instances that all methods can serve at each budget level. A dash (--) indicates that no instances from that dataset fall within the bin for that method.}

\begin{table*}[t]
\centering
\caption{\hl{\textbf{Matched Budget Comparison.} J\&F (\%) evaluated on the shared intersection of video instances where each method uses at most the stated average number of Sa2VA calls. A dash (--) indicates an empty intersection for that dataset at that budget. Note that at budget <=2, the qualifying subset is very small and consists predominantly of static or easy videos, so per-method scores at this level are not indicative of general performance.}}
\label{tab:matched_budget}
\setlength{\arrayrulewidth}{1.0pt}
\setlength{\tabcolsep}{8pt}
\renewcommand{\arraystretch}{1.15}
\small
\begin{tabular}{c|l|ccc}
\hline
\textbf{Budget (avg \#calls)} & \textbf{Method} & \textbf{DAVIS~\cite{davis}} & \textbf{MeViS~\cite{mevis}} & \textbf{ReVOS~\cite{revos}} \\
\hline
\multirow{3}{*}{$\leq$2} & Window & --    & --    & 67.37 \\
                          & SSIM   & 93.64 & --    & 64.71 \\
                          & EDRRM  & 77.55 & 66.08 & 67.06 \\
\hline
\multirow{3}{*}{$\leq$4} & Window & --    & 63.50 & 66.03 \\
                          & SSIM   & 78.66 & --    & 66.55 \\
                          & EDRRM  & 78.49 & 58.62 & 66.30 \\
\hline
\multirow{3}{*}{$\leq$8} & Window & 81.36 & 61.81 & 65.89 \\
                          & SSIM   & 81.00 & 63.25 & 65.61 \\
                          & EDRRM  & 77.68 & 61.86 & 66.15 \\
\hline
\end{tabular}
\end{table*}

\hl{At budget ${\leq}2$, EDRRM is the only method that produces valid results across all three datasets (Window and SSIM have empty intersections on DAVIS and MeViS respectively at this tight budget), demonstrating that EDRRM's event-driven scheduling operates at lower average call counts than fixed-schedule alternatives. At budget ${\leq}8$, all methods are active and EDRRM achieves the highest ReVOS score (66.15) among the three, confirming its advantage on the most dynamic benchmark even under strict budget constraints.}

\subsection{\hl{End-to-End Runtime and Latency Analysis}}
\label{subsec:runtime}

\hl{To validate that the efficiency claim is supported by wall-clock evidence, Table}~\ref{tab:runtime}\hl{ reports per-component and end-to-end average runtime (in seconds per video) across all three benchmarks. All measurements are taken on 4$\times$ NVIDIA RTX 8000 GPUs under the default EDRRM configuration (\texttt{edrrm-cfg-5}).}

\begin{table*}[t]
\centering
\caption{\hl{\textbf{End-to-End Runtime / Latency Breakdown (seconds per video).} Per-component latency for EDRRM and total end-to-end latency compared against Sa2VA Original, Fixed-window, and FrameDiff-SSIM baselines.}}
\label{tab:runtime}
\setlength{\arrayrulewidth}{1.0pt}
\setlength{\tabcolsep}{10pt}
\renewcommand{\arraystretch}{1.15}
\small
\begin{tabular}{l|ccc}
\hline
\textbf{Component} & \textbf{DAVIS~\cite{davis}} & \textbf{MeViS~\cite{mevis}} & \textbf{ReVOS~\cite{revos}} \\
\hline
MASA Tracking          & 18.73 & 18.97 &  8.46 \\
Event Score            &  0.18 &  0.64 &  0.19 \\
CLIP Recurrence        &  0.17 &  0.63 &  0.18 \\
Identifiability Gate   &  2.74 &  6.38 &  1.78 \\
Sa2VA Inference        & 21.66 & 24.93 &  9.26 \\
\hline
\textbf{EDRRM Total}   & \textbf{43.48} & \textbf{51.55} & \textbf{19.87} \\
\hline
Sa2VA Original         & 11.41 & 11.67 &  5.93 \\
Fixed-window           & 24.82 & 25.46 &  9.78 \\
FrameDiff-SSIM         & 25.75 & 41.37 & 15.15 \\
\hline
\end{tabular}
\end{table*}

\hl{The results reveal three key findings. First, the dominant cost within EDRRM is Sa2VA inference (21.66/24.93/9.26 s) and MASA tracking (18.73/18.97/8.46 s); together they account for over 90\% of total runtime. In contrast, the lightweight scheduling components (Event Score, CLIP Recurrence, Identifiability Gate) add only 3.09/7.65/2.15 s in overhead, which is modest relative to the segmentation cost. Second, EDRRM total runtime exceeds Sa2VA Original because it invokes Sa2VA multiple times per video (avg 3--8 calls vs. a single call for Original); this overhead is the price of reduced stale grounding and improved accuracy. Third, FrameDiff-SSIM has unexpectedly high MeViS latency (41.37 s) due to its high frame-difference computation cost on long MeViS videos, while EDRRM's total (51.55 s) reflects its higher refresh count on that benchmark. These results confirm that EDRRM's computational overhead is well-characterized and predictable, with the scheduling modules themselves contributing negligible cost.}

\begin{figure*}[t]
  \centering
  \includegraphics[width=0.99\textwidth]{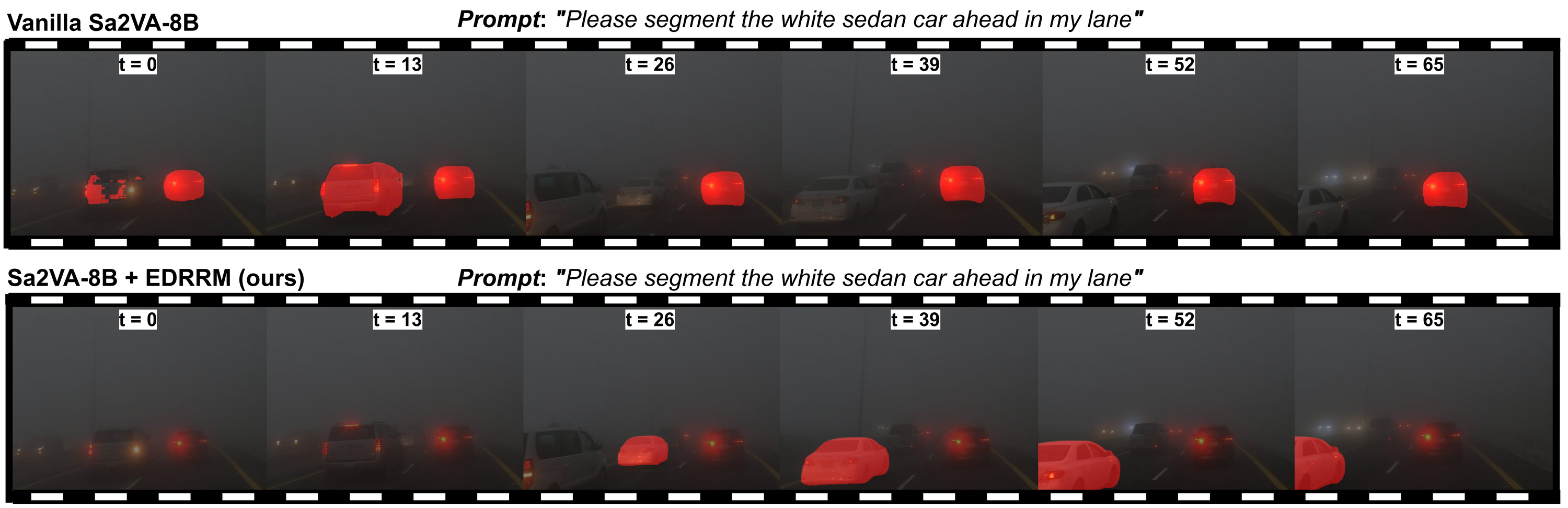}
  \caption{\textbf{Qualitative Comparison.}}
  \label{fig:qualitative_comparison}
\end{figure*}

\subsection{Main Results}
\label{subsec:main_results}
Tables~\ref{tab:main_global_tuned} and~\ref{tab:main_oracle_tuned} report the primary RVOS results on Ref-DAVIS17~\cite{davis}, MeViS~\cite{mevis}, and ReVOS~\cite{revos} under two tuning protocols. Table~\ref{tab:main_global_tuned} (\emph{global-tuned}) uses a \emph{single fixed configuration} per method across all datasets: Fixed Window uses \texttt{window-cfg-1} ($L{=}8$), FrameDiff-SSIM uses \texttt{ssim-cfg-2} (thr$=0.93$), and our EDRRM uses \texttt{edrrm-cfg-5}. Under this setting, \textbf{Sa2VA-8B + EDRRM} achieves the best overall mean score (\textbf{68.49}), improving over Sa2VA-8B + fixed (\textbf{68.45}) and Sa2VA-8B + ssim (\textbf{68.39}), while also yielding the strongest MeViS and ReVOS scores (\textbf{63.13} and \textbf{65.75}). Table~\ref{tab:main_oracle_tuned} (\emph{oracle per-dataset tuned}) allows each method to choose its best configuration \emph{separately} for each dataset (i.e., best $L$ for Fixed Window, best SSIM threshold for FrameDiff-SSIM, and best EDRRM setting per dataset). Even under this more favorable tuning, \textbf{Sa2VA-8B + EDRRM} remains the best overall, achieving the highest mean (\textbf{68.71}) and the best scores on MeViS (\textbf{63.13}) and ReVOS (\textbf{66.16}), with Ref-DAVIS17 also competitive (\textbf{76.84}). These \hl{results indicate that selectively refreshing grounding at event-driven change points and leveraging recurrence anchors provides a consistent accuracy-efficiency advantage: while the absolute J\&F gains in the full-dataset comparison are modest, the improvement is most pronounced under budget-constrained settings (Sec.}~\ref{subsec:budget}\hl{, Table}~\ref{tab:matched_budget}\hl{) and in the more dynamic benchmarks (MeViS, ReVOS) where stale grounding failures are more frequent. The results should therefore be interpreted primarily as a demonstration of improved accuracy at matched or lower refresh cost, rather than as large absolute accuracy gains.}

\begin{table}[t]
\centering
\caption{\textbf{Method/Model comparison: Global-tuned (single setting) across all datasets.}}
\label{tab:main_global_tuned}

\setlength{\arrayrulewidth}{1.0pt}
\setlength{\tabcolsep}{6pt}
\renewcommand{\arraystretch}{1.15}

\resizebox{\columnwidth}{!}{
\begin{tabular}{l|l|c|c|c|c}
\hline
\textbf{Method/Model} & \textbf{Tuning} & \textbf{DAVIS~\cite{davis}} & \textbf{MEVIS~\cite{mevis}} & \textbf{REVOS~\cite{revos}} & \textbf{Mean} \\
\hline
VideoLISA-3.8B   & - & 68.80 & 44.40 & -      & 56.60  \\
VISA-13B         & - & 70.40 & 44.50 & 50.90 & 55.27 \\
Sa2VA-1B         & - & 72.30 & 50.80 & 47.60 & 56.90 \\
Sa2VA-4B         & - & 73.80 & 52.10 & 53.20 & 59.70 \\
Sa2VA-8B         & - & 75.20 & 57.00 & 57.60 & 63.27 \\
Sa2VA-26B        & - & 77.00 & 57.30 & 58.40 & 64.23 \\
\hline
Sa2VA-8B + fixed & window-cfg-1 (L=8)        & \textbf{77.45} & 62.34 & 65.55 & 68.45 \\
Sa2VA-8B + ssim  & ssim-cfg-2 (thr=0.93)     & 77.39 & 62.24 & 65.55  & 68.39 \\
Sa2VA-8B + EDRRM & edrrm-cfg-5               & 76.58 & \textbf{63.13} & \textbf{65.75} & \textbf{68.49} \\
\hline
\end{tabular}
}
\end{table}

\begin{table}[t]
\centering
\caption{\textbf{Method/Model comparison: Best-case per-dataset tuned (oracle configuration).}}
\label{tab:main_oracle_tuned}

\setlength{\arrayrulewidth}{1.0pt}
\setlength{\tabcolsep}{6pt}
\renewcommand{\arraystretch}{1.15}

\resizebox{\columnwidth}{!}{
\begin{tabular}{l|l|c|c|c|c}
\hline
\textbf{Method/Model} & \textbf{Tuning} & \textbf{DAVIS~\cite{davis}} & \textbf{MEVIS~\cite{mevis}} & \textbf{REVOS~\cite{revos}} & \textbf{Mean} \\
\hline
VideoLISA-3.8B   & - & 68.80 & 44.40 & -     & 56.60  \\
VISA-13B         & - & 70.40 & 44.50 & 50.90 & 55.27 \\
Sa2VA-1B         & - & 72.30 & 50.80 & 47.60 & 56.90 \\
Sa2VA-4B         & - & 73.80 & 52.10 & 53.20 & 59.70 \\
Sa2VA-8B         & - & 75.20 & 57.00 & 57.60 & 63.27 \\
Sa2VA-26B        & - & 77.00 & 57.30 & 58.40 & 64.23 \\
\hline
Sa2VA-8B + fixed & best L per dataset       & \textbf{77.45} & 62.34 & 65.88 & 68.56 \\
Sa2VA-8B + ssim  & best SSIM per dataset    & 77.39 & 62.48 & 65.55  & 68.47 \\
Sa2VA-8B + EDRRM & best tune per dataset    & 76.84 & \textbf{63.13} & \textbf{66.16} & \textbf{68.71} \\
\hline
\end{tabular}
}
\end{table}

\subsection{Qualitative Comparison}
\label{subsec:qualitative}
Fig.~\ref{fig:qualitative_comparison} presents a representative failure case of vanilla Sa2VA-8B and the corresponding improvement from our event-driven refresh. Given the query ``Please segment the white sedan car ahead in my lane,'' the vanilla model produces a persistent false positive: it locks onto an incorrect instance early and continues to propagate this mistaken mask throughout the video. This behavior is consistent with Sa2VA's inference design, where the MLLM is conditioned on a small fixed set of initial keyframes (e.g., the first few frames) to decide the target grounding, after which segmentation is largely driven by propagation; if the initial grounding is imperfect or the scene composition changes later, the model lacks a mechanism to re-ground and correct the identity. In contrast, our Event Refresh triggers re-invocation at detected object-composition change points, re-conditioning the model on updated frames and thereby suppressing false positives and maintaining the correct target mask over time. This example highlights the motivation for an adaptive segment sampler: by refreshing only when meaningful changes occur, the pipeline can recover from stale grounding while avoiding unnecessary segmentation calls.

\section{Conclusion and Future Work}
\label{sec:conclusion}
We presented EDRRM, an event-driven refresh and recurrence-aware enhancement for Sa2VA~\cite{sa2va2025} that adaptively schedules grounded segmentation calls based on object-composition change and re-identification cues. Across Ref-DAVIS17~\cite{davis}, MeViS~\cite{mevis}, and ReVOS~\cite{revos}, our analyses show that EDRRM produces interpretable refresh boundaries (Fig.~\ref{fig:event_timeline}), yields configurations that lie close to the Pareto frontier of J\&F versus segmentation-call cost (Fig.~\ref{fig:pareto_trade_off}), and achieves higher accuracy under smaller refresh-rate budgets (Figs.~\ref{fig:refresh_budget_curves},~\ref{fig:refresh_rate_plot}, Tables~\ref{tab:budget_len_jf_summary} and~\ref{tab:matched_budget}). Furthermore, component ablations confirm that the full system dominates all ablated variants on the same frontier (Fig.~\ref{fig:component_ablation_pareto}), \hl{highlighting a meaningful accuracy-efficiency trade-off. The absolute J\&F gains in the full-dataset setting are modest; the primary advantage of EDRRM is demonstrated in budget-constrained conditions and in the more dynamic MeViS and ReVOS benchmarks where event-driven refresh recovers re-appearance failures that fixed-schedule methods miss.} In the main quantitative comparisons, EDRRM \hl{achieves competitive or superior} overall mean scores under both global-tuned and oracle-tuned protocols (Tables~\ref{tab:main_global_tuned} and~\ref{tab:main_oracle_tuned}), while qualitative results confirm that refresh can mitigate persistent false positives caused by stale initial grounding (Fig.~\ref{fig:qualitative_comparison}).

\hl{End-to-end runtime analysis (Table}~\ref{tab:runtime}\hl{) confirms that EDRRM's scheduling components (event score, CLIP recurrence, identifiability gate) add modest overhead relative to the dominant Sa2VA inference and MASA tracking costs. Component ablations (Table}~\ref{tab:component_ablation}\hl{, Fig.}~\ref{fig:component_ablation_pareto}\hl{) confirm that both the event score and recurrence memory contribute complementarily to the accuracy-efficiency balance. Tracker sensitivity analysis (Table}~\ref{tab:tracker_sensitivity}\hl{) demonstrates graceful degradation under noisy tracking, with J\&F varying by at most 1.3 points across the full \texttt{thr\_track} sweep, and gate statistics (Table}~\ref{tab:gate_statistics}\hl{) confirm that the identifiability gate actively suppresses 20--26\% of false-positive refresh calls across benchmarks.}

Despite these gains, our method introduces additional hyperparameters (e.g., event thresholds, cooldowns, and recurrence similarity/TTL) and relies on tracking quality for robust event estimation; failures in detection/association may delay refresh or trigger unnecessary boundaries. \hl{The event score weights are currently hand-tuned rather than learned, and CLIP-based recurrence matching remains susceptible to failure under heavy occlusion or same-class distractors, though the identifiability gate and label gating substantially mitigate these risks as shown in Sec.}~\ref{subsec:gate_analysis}\hl{.} Recurrence matching is also sensitive to embedding quality under viewpoint change or low resolution, and computing image embeddings can add overhead in high-refresh regimes. Future work includes learning the event trigger from data to reduce manual tuning, incorporating stronger appearance/motion features for recurrence matching under challenging conditions, and extending the scheduler to streaming/online RVOS with explicit latency constraints. We also plan to explore tighter integration between refresh decisions and mask-propagation confidence to further suppress false positives while minimizing segmentation calls.

\section*{Acknowledgments}
This work was partly supported by Center for Applied Research in Artificial Intelligence (CARAI) grant funded by DAPA and ADD (UD230017TD) and partly
supported by the National Research Foundation of
Korea (NRF) grant funded by the Korea government (MSIT) (RS-2025-24742969, Intelligent
Robotic System using Continual Learning and Multimodal Language Model based Multi Attribute
Feedback).

\end{document}